\documentclass{article}
\usepackage{graphicx}
\usepackage{amssymb} 
\usepackage{xcolor}
\usepackage{amsmath}
\usepackage[margin=1in]{geometry}
\usepackage{booktabs}
\usepackage{longtable}
\usepackage{tabularx}
\usepackage{array}
\usepackage{enumitem}
\usepackage{array}
\usepackage{float}
\usepackage{fourier}
\usepackage{microtype}
\usepackage[font=small,labelfont=bf,labelsep=period]{caption}
\usepackage{siunitx}
\DeclareSIUnit{\angstrom}{\textup{\AA}}
\usepackage[style=authoryear,maxcitenames=2,maxbibnames=30,natbib=true,eprint=false]{biblatex}
\AtEveryBibitem{\clearfield{issn}}
\usepackage{placeins}
\usepackage[version=4]{mhchem}
\usepackage{algorithm}
\usepackage{algpseudocode}
\usepackage{listings}

\definecolor{lstbg}{gray}{0.93}
\usepackage{hyperref}
\usepackage{xparse}

\definecolor{clrprompt}{HTML}{8B0000}
\definecolor{clrthink}{HTML}{1a5276}
\definecolor{clrreason}{HTML}{6C3483}
\definecolor{clranswer}{HTML}{117A65}
\newcommand{\tprompt}[1]{{\small\ttfamily\bfseries\textcolor{clrprompt}{#1}}}
\newcommand{\tthink}[1]{{\small\ttfamily\bfseries\textcolor{clrthink}{#1}}}
\newcommand{\treason}[1]{{\small\ttfamily\bfseries\textcolor{clrreason}{#1}}}
\newcommand{\tanswer}[1]{{\small\ttfamily\bfseries\textcolor{clranswer}{#1}}}
\definecolor{clrlabel}{HTML}{B45F06}
\newcommand{\tlabel}[1]{{\small\ttfamily\bfseries\textcolor{clrlabel}{#1}}}

\definecolor{citegray}{gray}{0.25} 

\newcommand{\mycitestyle}[1]{\textcolor{citegray}{\small #1}}

\NewCommandCopy{\oldcite}{\cite}
\RenewDocumentCommand{\cite}{o o m}{%
  \mycitestyle{%
    \IfNoValueTF{#2}{\oldcite[#1]{#3}}{\oldcite[#1][#2]{#3}}%
  }%
}

\NewCommandCopy{\oldcitep}{\citep}
\RenewDocumentCommand{\citep}{o o m}{%
  \mycitestyle{%
    \IfNoValueTF{#1}{%
      \oldcitep{#3}
    }{%
      \IfNoValueTF{#2}{%
        \oldcitep[#1]{#3}
      }{%
        \oldcitep[#1][#2]{#3}
      }%
    }%
  }%
}

\NewCommandCopy{\oldcitet}{\citet}
\RenewDocumentCommand{\citet}{o o m}{%
  \mycitestyle{%
    \IfNoValueTF{#1}{%
      \oldcitet{#3}%
    }{%
      \IfNoValueTF{#2}{%
        \oldcitet[#1]{#3}%
      }{%
        \oldcitet[#1][#2]{#3}%
      }%
    }%
  }%
}

\NewCommandCopy{\oldcitealp}{\citealp}
\DeclareRobustCommand{\citealp}[2][]{%
  \mycitestyle{\oldcitealp[#1]{#2}}%
}

\newsavebox{\mycaptionbox}  

\title{Evaluating the Progression of Large Language Model Capabilities for Small-Molecule Drug Design}
\author{
    Shriram Chennakesavalu\thanks{Corresponding author: \texttt{chennakesavalu.shriram@gene.com}},
    Kirill Shmilovich\thanks{These authors contributed equally to this work.},
    Hayley Weir\footnotemark[2],
    Colin Grambow\footnotemark[2],\\
    John Bradshaw,
    Patricia Suriana,
    Chen Cheng,
    Kangway Chuang\\
    {\small Prescient Design, Genentech, South San Francisco, CA, USA}
}
\date{\today}

\begin{document}

\maketitle
\begin{abstract}%
{\noindent
Large Language Models (LLMs) have the potential to accelerate small molecule drug design due to their ability to reason about information from diverse sources and formats. However, their practical utility remains unclear due to the lack of benchmarks that reflect real-world scenarios. In this work, we introduce a suite of chemically-grounded tasks spanning molecular property prediction, molecular representation transformations, and molecular design. Importantly, we formulate these tasks as reinforcement learning (RL) environments, enabling a unified approach for evaluation and post-training. Across three model families, we find that frontier models are increasingly proficient at chemical tasks, but that there is significant room for improvement, especially in experimental settings with low data. Critically, we show that RL-based post-training can substantially improve performance. A smaller model post-trained on our environments becomes competitive with state-of-the-art frontier models, despite a significantly weaker base model. This suggests a practical route toward employing LLMs in drug discovery; by combining carefully-designed evaluation tasks with targeted post-training, we can both elucidate and close critical capability gaps.}
\end{abstract}

\section{Introduction}
\label{sec:introduction}

Drug-discovery campaigns are truly multidisciplinary efforts that rely on a vast set of computational, experimental, and clinical approaches to successfully bring a drug to market~\citep{paul2010how-a99, hughes2011principles-6a9, macalino2015role-4e0, rcz2025changing-be8}. 
Designing generalist systems that can seamlessly synthesize information and leverage tools across the full drug discovery pipeline can potentially reduce the time and cost of designing a drug and help us avoid failure modes that plague many campaigns today (e.g., clinical inefficacy, safety issues)~\citep{harrison2016phase-015,paul2010how-a99}. 

Recently, Large Language Models (LLMs) have emerged as truly generalist models, capable of synthesizing knowledge and generating new ideas across many diverse disciplines~\citep{anthropic2026claude-8ac,openai2023gpt-4-081,google2025gemini-fa4}. In many real-world settings, LLMs are often augmented with tools, enabling these so-called agents to carry out more sophisticated tasks, via information acquisition from and/or modification of an external environment~\citep{yao2022react-fe3, bran2024augmenting-805, boiko2023autonomous-ead}. LLM-based agents have recently been deployed in many drug discovery contexts, including for target identification, lead optimization, and toxicity prediction~\citep{bran2024augmenting-805,boiko2023autonomous-ead,ramos2024review-bfa}. While these are compelling demonstrations, many of these agents are limited by the performance of the base LLM~\citep{bran2024augmenting-805, putta2024agent-571}. Frontier models struggle at many of the fundamental biological and chemical tasks \citep{narayanan2025training-64a, anthropic2026claude-8ac} that underpin many of these drug discovery agents, and building truly robust agents for drug discovery will require stronger base models.

Understanding both the current capabilities of existing LLMs and the expected capabilities of future LLMs is crucial to designing agents for drug discovery. While frontier LLMs have known limitations for drug discovery tasks~\citep{narayanan2025training-64a}, these models are rapidly advancing and this progression will ultimately inform efforts to build agents, including identifying realistic use-cases and understanding how to allocate resources between model training and harness engineering. Understanding this progression in capabilities of models is challenging, in large part, due to the closed nature of many of these models. Both open-weight and closed frontier models generally do not disclose their training recipes, making it difficult to forecast how models will evolve in the short and long-term.

Early generations of LLMs were primarily improved through the scaling of pretraining, by increasing model size, compute, and dataset size~\citep{kaplan2020scaling-1ce}. This approach enabled gains to be realized across tasks \textit{generally}. More recent models have been improved by extensive post-training, which has resulted in a so-called ``jagged frontier,'' where model capabilities in certain domains far outstrip capabilities in other domains~\citep{deepseek-ai2025deepseek-r1-54c, team2025kimi-0f3, yang2025qwen3-84a, narayanan2025training-64a}. Current training recipes of LLMs can be decomposed into two parts: base-model training and post-training~\citep{dubey2024llama-1fa, nvidia2025nvidia-473, olmo2025olmo-1b9, ren2024learning-36f}. Base-model training, which includes pretraining on vast datasets and midtraining on large but higher-quality datasets, fundamentally increase the knowledge and skills of the model~\citep{olmo2025olmo-1b9}. Post-training, such as with Reinforcement Learning (RL) methods like Direct Preference Optimization (DPO)~\citep{rafailov2023direct-689} and Group Relative Policy Optimization (GRPO)~\citep{deepseek-ai2025deepseek-r1-54c, shao2024deepseekmath-82a} sharpen the capabilities of the model on particular skills acquired during base-model training. 

With base-model training already incorporating knowledge from diverse corpora, one perspective for evaluating the potential capabilities of frontier models is to assess how easily their performance can be ``sharpened'' for a domain of interest using a post-training-only training scheme~\citep{yue2025does-b33}. For open-weight models, we can directly assess this potential by measuring the gap between the base model and a post-trained model. For closed models, we assess their potential capabilities by comparing the performance of subsequent iterations of a model within a model family. While the differences between various model iterations are not publicly transparent, examining the differences in model performance over time can serve as a proxy for how easily the model can be improved.

In this paper, we focus on a narrow drug discovery setting---small-molecule design---and examine the capabilities and trainability of three families of frontier models: GPT-5, Claude Opus~4, and Qwen-30B-A3B. We present a collection of tasks, ranging from property prediction to chemical language translation to molecular design. Importantly, we frame these tasks as Reinforcement Learning (RL) environments, enabling us to also post-train open-weight models on these tasks. We analyze the progression of model capabilities across model iterations within each of these families and find that generally subsequent iterations of models perform better on these tasks. Additionally, we find that our post-trained model is competitive with closed frontier models on a multi-turn molecular design task that simulates real-world lead optimization despite the base model being significantly weaker. Lastly, we find that frontier models and post-trained models struggle on experimental datasets with limited data suggesting that many of these tasks are out-of-distribution for LLMs currently and that we require further base-model training before LLMs can be meaningfully useful for these tasks.

\section{Related Work}
\label{sec:related-work}

\paragraph{Benchmarking LLMs.}
Benchmarking and evaluation have been a key driver of modern ML advancements across multiple application areas (\citealp{deng2009imagenet-dd3,rajpurkar2016squad-638,huang2021therapeutics-d92}\mycitestyle{;} \citealp[Chp.8]{hardt2022patterns-873}), although such practices are not without their issues~\citep{dehghani2021benchmark-944,bowman2021what-c39}. 
As much of the development of LLMs has shifted into private settings, new releases of LLMs are often first evaluated and judged by their performance on a commonly used set of benchmarks (see, e.g., Table 2 of \citet{openai2023gpt-4-081}, or Section 2 of \citet{anthropic2026claude-8ac}).
These benchmarks cover a wide range of problem domains such as coding~\citep{jimenez2023swe-bench-947}, math~\citep{cobbe2021training-245}, tool-use~\citep{bandi2026mcp-atlas-e3a}, and others~\citep{hendrycks2020measuring-d9c, Phan2025}.

\paragraph{Benchmarking LLMs on scientific tasks.}
The performance of LLMs has also been investigated in a number of scientific contexts.
These studies range from studying relatively simple one-step tasks, such as computational molecular design or understanding~\citep{Liu2024,ramos2024review-bfa,cai2025mollangbench-f1c,lu2024moleculeqa-75f,huang2024chemeval-192,runcie2026assessing-7e2,mirza2024are-8e3,nascimento2023do-748}, to more complex, often multi-turn jobs, including knowledge synthesis from scientific literature~\citep{patel2025deepscholar-bench-b4c} and conducting or proposing wet-lab experiments~\citep{xu2025probing-ff6,bran2024augmenting-805,gottweis2025towards-2a8,boiko2023autonomous-ead}. 
While valuable, these efforts can be limited in scope or rely on incorporating external information to guide the responses of the LLM (e.g., via external tools or other ``personas'' of the same base LLM). 
Furthermore, these investigations demonstrate that the performance of the base model is generally lacking for fundamental chemical and biological tasks, motivating the need for training models that can be more robust in real-world settings.

\paragraph{Training LLMs for scientific tasks.}
Recent efforts have aimed at training reasoning LLMs to improve performance on scientfic tasks. These efforts have involved incorporating additional specialized encoders or exploring the use of RL-based post-training to adapt the model to problems of interest~\citep{jablonka2024leveraging-530,istrate2026rbio1-2fb,fallahpour2025bioreason-3ce,narayanan2025training-64a,prabhakar2025omniscience-ccc}. These models are often characterized by their ability to reason using long chain-of-thought (CoT) and approach a task with deliberate, system-2-style thinking~\citep{wei2022chain-of-thought-6e3}.
For example, the \texttt{ether0} effort~\citep{narayanan2025training-64a} distilled reasoning-like ability into a non-reasoning base model from a stronger reasoning teacher model and carried out RL-based training to improve the performance of the model on a collection of chemistry tasks; however, they do not look at how this added capability can help more complex, multi-step molecular optimization routines. Other recent work, carried out by \citet{fallahpour2025bioreason-3ce}, used a base reasoning model and used custom encoders to encode biological modalities into the model. That said, they used a comparatively small model (4B parameters) and it is unclear whether specialized encoders are required for medium-scale (10--100B) and large-scale (100B--1T+) models.

\section{Background}
\label{sec:background}

Consider a trained LLM represented by a reference policy $\pi_{\textrm{ref}}: \mathcal{X} \times \mathcal{Y}\rightarrow[0, 1]$, from which we can sample a response $\boldsymbol{y} \in \mathcal{Y}$ given a prompt $\boldsymbol{x} \in \mathcal{X}$, where $\mathcal{X}$ and $\mathcal{Y}$ denote discrete spaces. We are interested in employing a policy for a collection of tasks $\mathcal{T} = \{\tau_1, \tau_2, \dots, \tau_T\}$, where each task is defined as $\tau_i = \{\mathcal{D}_i, R_i\}$. Here, $\mathcal{D}_i = \{\boldsymbol{x}_1^{(i)}, \boldsymbol{x}_2^{(i)}, \dots, \boldsymbol{x}_{n_i}^{(i)}\}$ denotes the dataset over $\mathcal{X}$ specific to the $i$-th task, and $R_i: \mathcal{X} \times \mathcal{Y}\rightarrow\mathbb{R}$ is the corresponding task-specific reward function that evaluates the quality of a response $\boldsymbol{y}$ given a prompt $\boldsymbol{x}$. We treat $\mathcal{D}_i$ as inducing an empirical distribution over prompts and write $\boldsymbol{x} \sim \mathcal{D}_i$ to denote uniform sampling from this dataset.

Ultimately, we are interested in finding an optimal policy $\pi^*$ that maximizes the expected reward across all tasks:
\begin{equation}
\pi^* = \arg\max_{\pi}
\frac{1}{T} \sum_{i=1}^T
\mathbb{E}_{\boldsymbol{x} \sim \mathcal{D}_i,\; \boldsymbol{y} \sim \pi(\cdot \mid \boldsymbol{x})}
\left[
R_i(\boldsymbol{x}, \boldsymbol{y})
\right]
\label{eq:max_rewards}
\end{equation}
There are numerous possible strategies for designing an optimal policy that involve some combination of designing an external scaffold around the LLM (e.g.,\ prompt engineering, harness engineering) and/or training the internal weights of the LLM (e.g.,\ supervised fine-tuning, reinforcement learning). In this work, we exclusively focus on RL-based post-training of a base model and leave exploration of other methods to future work.

\subsection{Reinforcement Learning}
\label{sec:rl}
We carry out on-policy reinforcement learning using a variant of Group Relative Policy Optimization (GRPO)~\citep{shao2024deepseekmath-82a} to learn a policy $\pi_{\theta}$ that optimizes Eq.~\eqref{eq:max_rewards}. GRPO-like algorithms are desirable for their computational and practical simplicity; they do not require training a separate value function and instead rely on computing a normalized advantage function. Specifically, for a given prompt $\boldsymbol{x} \sim \mathcal{D}_i$, we sample a group of $G$ responses $\{\boldsymbol{y}_j\}_{j=1}^G \sim \pi_\theta(\cdot \mid \boldsymbol{x})$ from the current policy $\pi_\theta$. The normalized advantage $\hat{A}(\boldsymbol{x}, \boldsymbol{y}_j)$ for a specific response $\boldsymbol{y}_j$ is computed by standardizing its reward relative to the group:
\begin{equation}
    \hat{A}(\boldsymbol{x}, \boldsymbol{y}_j) = \frac{R_i(\boldsymbol{x}, \boldsymbol{y}_j) - \mu_G}{\sigma_G}
\end{equation}
where $\mu_G$ and $\sigma_G$ denote the mean and standard deviation of the rewards for the sampled group.

In this work, we use the DAPO-variant of GRPO~\citep{yu2025dapo-2df}
\begin{equation}
\mathcal{J}(\theta)
=
\frac{1}{T} \sum_{i=1}^T
\mathbb{E}_{\boldsymbol{x} \sim \mathcal{D}_i}
\left[
\left.
\frac{1}{\sum_{j=1}^G |\boldsymbol{y}_j|}
\sum_{j=1}^G
\sum_{t=1}^{|\boldsymbol{y}_j|}
\textrm{clip}
\left(
\frac{\pi_\theta(y_{j,t} \mid \boldsymbol{x}, \boldsymbol{y}_{j,<t})}
{\pi_{\textrm{old}}(y_{j,t} \mid \boldsymbol{x}, \boldsymbol{y}_{j,<t})},
\hat{A}_j,
\epsilon_{\textrm{low}},
\epsilon_{\textrm{high}}
\right)
\;\right|_{\{\boldsymbol{y}_j\}_{j=1}^G \sim \pi_{\textrm{old}}}
\right]
\label{eq:dapo}
\end{equation}
In the purely on-policy training setting, where samples are drawn from the current policy $\pi_\theta$, the gradient of the objective reduces to a REINFORCE-like policy gradient~\citep{williams1992simple-f0c} with sequence-level advantage
\begin{equation}
\nabla_\theta \mathcal{J}(\theta)
=
\frac{1}{T} \sum_{i=1}^T
\mathbb{E}_{\boldsymbol{x}\sim\mathcal{D}_i,\;\{\boldsymbol{y}_j\}_{j=1}^G\sim\pi_\theta}
\left[
\frac{1}{\sum_{j=1}^G |\boldsymbol{y}_j|}
\sum_{j=1}^G
\hat A_j \,
\nabla_\theta \log \pi_\theta(\boldsymbol{y}_j\mid \boldsymbol{x})
\right].
\label{eq:grad_J}
\end{equation}
Eq.~\eqref{eq:grad_J} provides a policy gradient estimator~\citep{williams1992simple-f0c, NIPS1999_464d828b} for Eq.~\eqref{eq:max_rewards}, with implicit length normalization.

\section{Small-Molecule Tasks}
\label{sec:tasks}

We curated a diverse set of chemistry tasks that reflected key aspects of small-molecule drug discovery, including molecular understanding, molecular property prediction, and molecular design. Specifically, the set of tasks include prediction of RDKit properties, prediction of molecular potency and DMPK properties on internal and external datasets~\citep{2021fs_mol}, translation between different molecular representations, and generation of molecules given property constraints. Collectively, these tasks are constructed in such a way evaluate a model's ability to meaningfully reason about small-molecule chemistry in real-world settings. The task suite is organized into six broad groups:
\begin{itemize}
    \item \textbf{RDKit property prediction}: Given a molecule and an \texttt{RDKit} property, the model must predict the property value for that molecule. Because these are general properties (e.g., molecular weight), we provide no additional context about what the property is. We use a reward model that scales the prediction error between 0 and 1 (Eq.~\ref{eq:exp-mse}).
    \item \textbf{Experimental prediction}: Given an in-context collection of molecules with their measured property, the model must predict the property of a held out molecule. Because these are target and assay-specific properties, we include molecules in the context to provide more information about the property of interest in order to guide prediction. To ensure that the in-context molecules are useful for predicting the property of the query molecule, we carry out a clustering procedure using HDBSCAN~\citep{scikit-learn} based on the Dice similarity of the maximum common substructure (MCS). We then design prompts such that all in-context molecules and the query molecule come from the same cluster. This group of tasks include prediction on internal potency and DMPK datasets and the FS-Mol dataset~\citep{2021fs_mol}. Here, we use the same reward model as in the previous group, which scales the prediction error between 0 and 1 (Eq.~\ref{eq:exp-mse}).
    \item \textbf{Multiple choice}: We reformulate the tasks in the previous group as a multiple-choice question, where we provide the same in-context molecules and ask the model which of four molecules has the highest/lowest property. Additionally, we include a task consisting of a query molecule in SMILES representation and four molecules, where one of the molecules is the same as the query but represented by a different SMILES and the other three molecules are different but have a SMILES similar to the query. The model is prompted to identify which molecule is the same as the query molecule. All rewards for this task are a binary reward on the correctness (Eq.~\ref{eq:binary-equivalence}).
    \item \textbf{Transformation}: This group of tasks consists of translating between various molecular representations (SMILES, IUPAC, Molecular Formula, Dominant Tautomer, Dominant Protomer, Murcko Scaffold). The reward function consists of a weighted sum of a binary correctness reward and a similarity reward that computes how similar the predicted molecule is to the ground truth answer (Eq.~\ref{eq:equivalence}). This reward provides a denser signal for the model to learn the task.
    \item \textbf{Multiproperty constrained generation}: In this task, we provide a set of constraints (ranging from 1 to 5 regular constraints and up to 3 additional element-count constraints) and prompt the model to generate a molecule satisfying all those constraints. The regular constraints include physicochemical, DMPK, and molecular scaffold constraints. The reward for this task is the fraction of constraints satisfied (Eq.~\ref{eq:constraint-sat}).
    \item \textbf{Other}: We include additional miscellaneous tasks, consisting of substructure classification, reaction outcome prediction, and MCS-identification tasks.
\end{itemize}
Table~\ref{tab:aspen_task_summary} summarizes the major groups and the datasets used to construct these tasks. For further details on the tasks, evaluation criteria, and reward functions see Appendix ~\ref{app:task-details-appendix}.

\begin{table}[htb]
\centering
\footnotesize
\renewcommand{\arraystretch}{1.2}
\caption{Summary of task groups. Reward details can be found in Appendix~\ref{sec:rewards}.}
\label{tab:aspen_task_summary}
\begin{tabularx}{\textwidth}{lXX}
\toprule
\textbf{Task group} & \textbf{Input $\rightarrow$ Output} & \textbf{Data} \\
\midrule
RDKit prediction &
SMILES $\rightarrow$ numeric value &
ZINC (100{,}000 molecules)~\citep{tingle2023zinc-22-e9c} \\

Experimental prediction &
In-context set + query molecule $\rightarrow$ numeric value &
Internal potency + DMPK data; FS-Mol~\citep{2021fs_mol}  \\

Transformation &
Molecular representation $\rightarrow$ Molecular representation &
ZINC (100{,}000 molecules)~\citep{tingle2023zinc-22-e9c} \\

Multiprop generation &
Constraints $\rightarrow$ SMILES &
ZINC (100{,}000 molecules)~\citep{tingle2023zinc-22-e9c} \\

\bottomrule
\end{tabularx}
\end{table}

\section{Results}
\label{sec:training}

\subsection{Post-training a 30B parameter LLM}
\label{sec:post-training}

\begin{figure}[htb]
    \centering
    \includegraphics[width=\textwidth]{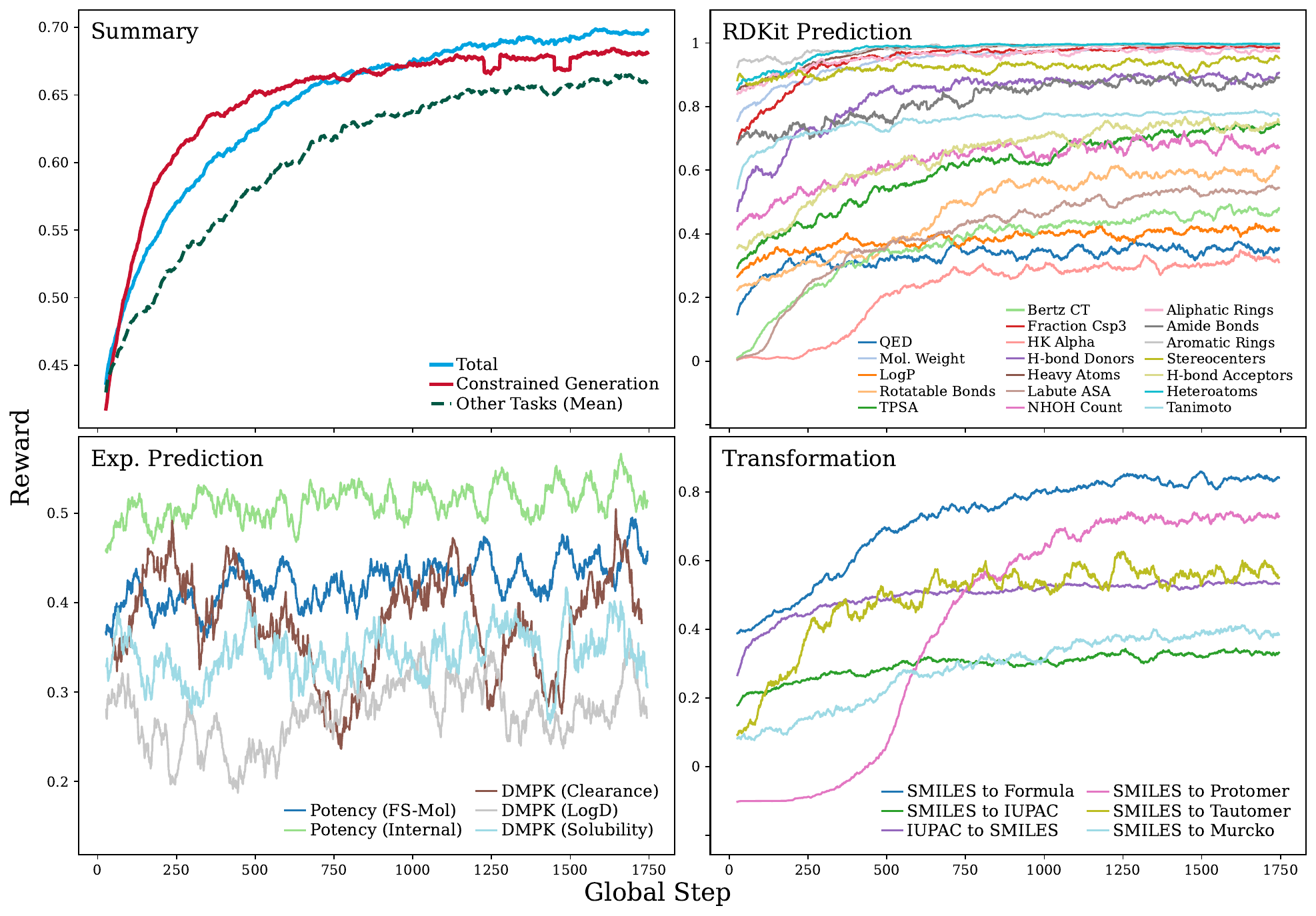}
    \caption{Reward trajectories over global step during one epoch of RL post-training of \texttt{Qwen3-30B-A3B-Thinking-2507}. Total reward rises steadily and begins to plateau, with especially strong gains in constrained generation, which is effectively oversampled relative to the other tasks because it contributes far more prompts ($\sim$300k vs.\ at most $\sim$20k per other task). Many RDKit and transformation tasks improve---often with sigmoidal learning curves---whereas experimental property prediction and the hardest representation-translation tasks remain noisy or improve only modestly.}
    \label{fig:reward-curves}
\end{figure}
We train \textbf{Aspen} via RL-based post-training of \texttt{Qwen3-30B-A3B-Thinking-2507}~\citep{yang2025qwen3-84a}. The model is a Mixture-of-Experts (MoE) language model with 30B total parameters, with 10 total experts, each consisting of 3B parameters on the tasks detailed in Section~\ref{sec:tasks}. For the post-training experiments here, we choose to forego any Supervised Fine-Tuning (SFT), largely because the base model was able to generate traces with reasoning structure. Additionally, we use \texttt{GRPOTrainer} from the HuggingFace \texttt{trl}~\citep{vonwerra2020trl} package to carry out all training using the DAPO loss (Eq.~\ref{eq:dapo}) due to its simplicity. Our experiments carried out RL synchronously (i.e., inference and training steps occur sequentially). We chose a synchronous approach due to the maturity of \texttt{GRPOTrainer} at the time of running the experiments and because all of our training environments were single-turn and did not incorporate additional tools. See Appendix~\ref{app:training_details} for all further training details.

We run training for a single epoch and plot the reward over step across all tasks and within each task in Figure~\ref{fig:reward-curves}.
We note that there is a deliberate task imbalance, where the molecular generation task (Constrained Generation) had around 300k prompts, while all other tasks had at most 20k prompts. 
Some tasks had less than 20k prompts due to small dataset sizes and strict filtering criteria. From Figure~\ref{fig:reward-curves}, we see that the overall reward is steadily increasing throughout the epoch and appears to plateau near the end.
We also observe sigmoidal-like learning for many tasks (e.g., SMILES to Protomer), where the performance on the task is very low for a while, rapidly rises, and then slowly plateaus afterwards.
In other tasks, the learning is more immediate (e.g., molecular weight prediction), while in difficult tasks, performance is stagnant throughout training. We see that the model generally struggles with experimental property prediction and translating between SMILES and IUPAC representations.
Ultimately, RL serves to extract and sharpen latent knowledge that exists in the base model. For many of the tasks here, we see that RL on its own---with some prior reward engineering---can quickly improve the performance of a model on that task.
However, for many of the tasks we struggle with (e.g., potency prediction), it is likely that the knowledge underlying these tasks is out-of-distribution and no amount of RL training will measurably improve the performance as there is no underlying knowledge to ``sharpen''.
Instead, we will require a training procedure (e.g., midtraining) that injects new knowledge into the base model before any final RL training. We leave exploration of other training procedures to future work.

\subsection{Single-turn environments}
\label{sec:single-turn}

\begin{figure}[p]
    \centering
    
    \newcommand{\lollipopcaptiontext}{Comparison of how model families (columns) are improving across our suite of tasks (rows). Within each group, tasks are sorted by difficulty (judged by average model performance), * denotes internal tasks (i.e., using our proprietary experimental data), and $\dagger$ denotes tasks that Aspen is not trained on (but are included for a more comprehensive assessment). Points out of range are marked with $\blacktriangleleft$.
    Property-prediction tasks are measured in terms of $R^2$, multiple choice is accuracy, transformation reports the fraction of outputs that are chemically equivalent to the ground-truth structure, multiproperty-constrained generation reports valid response and constraint satisfaction rates, substructure is measured by accuracy, and reaction outcome and MCS by chemical equivalence.
    While older models often struggle, even on simple tasks, this is beginning to change with the latest generation of models, and further improvement is often possible with careful post-training routines (especially in smaller open models with worse starting knowledge).}
    
    \sbox{\mycaptionbox}{\parbox{\textwidth}{\small\textbf{Figure X. } \lollipopcaptiontext}}
    
    \includegraphics[
        width=\textwidth,
        height=\dimexpr\textheight-\ht\mycaptionbox-\dp\mycaptionbox-\abovecaptionskip\relax,
        keepaspectratio
    ]{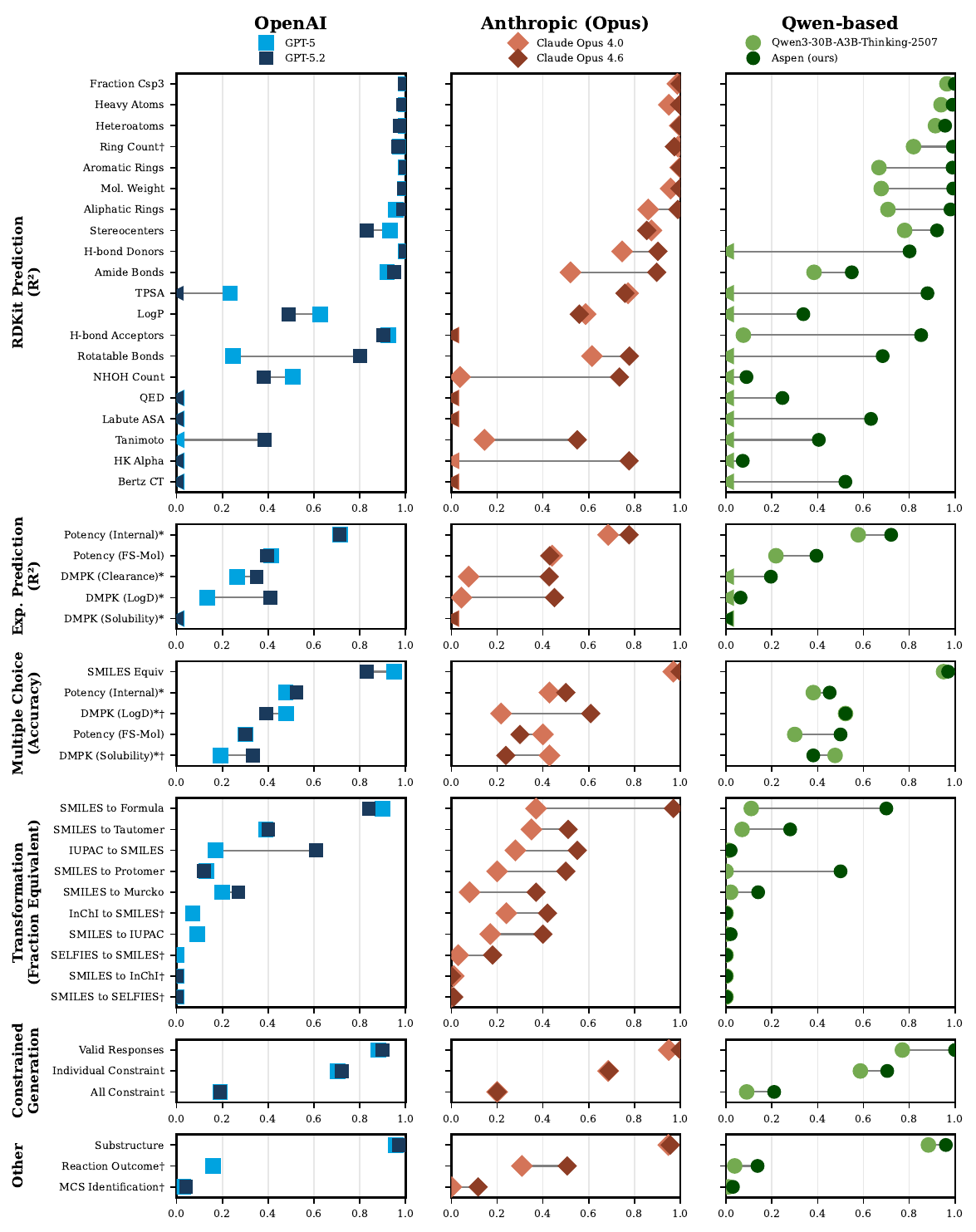}
    
    \caption{\lollipopcaptiontext}
    \label{fig:pairedLollipop}
\end{figure}

Figure~\ref{fig:pairedLollipop} summarizes model performance across our suite of single-turn tasks. The figure is meant to support two complementary comparisons. First, for cross-family benchmarking, the most relevant closed-model reference points for the Qwen/Aspen comparison are GPT-5 and Claude Opus~4.0, as these were released around the same time frame when \texttt{Qwen3-30B-A3B-Thinking-2507} became available. Second, and more importantly, the figure shows how capability evolves within each family over time: GPT-5 to GPT-5.2, Claude Opus~4.0 to Claude Opus~4.6, and Qwen to Aspen. Viewed this way, the dominant pattern is that later variants tend to improve on the harder chemical tasks, although the strength of that trend differs substantially by family. The improvement is strongest for Anthropic, suggesting that a substantial amount of effort has been devoted to training on chemical tasks in the recent model releases~\citep{anthropic2026claude-8ac}). We see a similar improvement between the base model and Aspen, where the post-trained model outperforms the base model on all trained tasks. Finally, the OpenAI pair shows a more heterogeneous shift, with clear gains on some difficult tasks but not a monotonic improvement on every single-turn evaluation.

\paragraph{RDKit property prediction tasks.} The RDKit prediction row is particularly revealing because it separates near-saturated descriptors from harder derived quantities. All model families are already strong on relatively simple, mostly counting-like properties such as the fraction of \textit{sp3} carbons, heavy-atom count, heteroatom count, ring count, aromatic/aliphatic-ring count, and molecular weight, all of which can be determined from relatively direct topological or compositional tallies.

By contrast, H-bond-donor/acceptor count, NH/OH count, amide-bond count, and to a lesser extent rotatable-bond count are still count-valued outputs, but they require chemically typed counting: the model has to recognize local functional-group identity, valence, aromaticity, protonation state, and in some cases tautomeric context, rather than merely tally atoms or tokens from the input SMILES string. This likely explains why the base Qwen model can be reasonably good on heavy atoms and heteroatoms ($R^2 = 0.94$ and 0.91) while performing very poorly on H-bond donors/acceptors, rotatable bonds, and NH/OH count (\num{-0.20}, 0.08, \num{-1.29}, and \num{-0.44}, respectively). Similarly, TPSA, LogP, QED, Labute ASA, Tanimoto similarity, Hall-Kier alpha, and Bertz CT are more complex derived properties that require better reasoning over the SMILES to correctly determine molecular connectivity.

Aspen fixes much of this---raising H-bond donors to 0.80, H-bond acceptors to 0.85, rotatable bonds to 0.68, and TPSA to 0.88, but NH/OH count remains low at 0.09. Despite its apparent simplicity, NH/OH count depends on implicit hydrogens and local chemical context, making it difficult to determine even when the model has become broadly better at local functional-group reasoning. Additionally, it is often low-range, so modest off-by-one errors on a relatively small number of molecules can sharply depress $R^2$. The fact that these donor/acceptor-style counts remain uneven across families suggests that they are not trivial bookkeeping tasks, but rather compact tests of whether the model has learned chemically correct local semantics.

Claude Opus~4.6 also improves markedly over Opus~4.0 on several hard RDKit targets, especially Hall-Kier alpha, NH/OH count, and Tanimoto similarity. The OpenAI family improves more selectively: GPT-5.2 is much better on rotatable bonds and Tanimoto similarity, but regresses on descriptors such as TPSA and QED.

\paragraph{Experimental property prediction and multiple choice tasks.}

Experimental property prediction and the related multiple-choice potency/DMPK tasks probe a different regime beyond deterministic graph-derived properties: transfer of medicinal-chemistry intuition to sparse experimental data. Here Aspen’s improvement over Qwen on proprietary internal potency prediction is substantial (0.58$\rightarrow$0.72), and the gains on DMPK clearance and DMPK LogD move Aspen from negative to positive $R^2$ (\num{-0.19}$\rightarrow$0.20 and \num{-0.30}$\rightarrow$0.06). The multiple-choice results tell a similar but noisier story. Aspen improves on the potency questions, from 0.30 to 0.50 on FS-Mol and from 0.38 to 0.45 on the internal potency set, while remaining flat on DMPK LogD and slightly worse on DMPK solubility (0.48$\rightarrow$0.38), although Aspen was not explicitly trained on the latter two tasks.

Across families, the broader lesson is that later models are clearly getting better at chemistry grounded in experimental data, with the recent boost in performance of Opus~4.6 on DMPK properties being particularly notable, but performance is worse and progress is much less monotonic than on the easiest RDKit descriptors. For example, DMPK solubility prediction is not solved by any model: all models still have negative $R^2$ on that task. This poor performance across experimental tasks is likely because they depend on higher-level medicinal-chemistry generalization rather than deterministic structure-to-property rules.

\paragraph{Molecular representation transformation tasks.}

Tasks requiring transformations between molecular representations deserve separate emphasis because they require exact structure-preserving conversion between chemical languages. Aspen improves sharply over Qwen on several transformations---SMILES$\rightarrow$formula (0.11$\rightarrow$0.70), tautomer generation (0.07$\rightarrow$0.28), protomer generation (0.00$\rightarrow$0.50), and Murcko scaffold extraction (0.02$\rightarrow$0.14)---but still remains near zero accuracy on the hardest nomenclature and representation tasks, including IUPAC$\rightarrow$SMILES (0.01$\rightarrow$0.02), SMILES$\rightarrow$IUPAC (0.00$\rightarrow$0.02), as well as conversion of SMILES to/from InChI/SELFIES (although these were not explicitly trained on).

The gap in SMILES$\rightarrow$IUPAC performance is especially interesting because Aspen was trained with a dense IUPAC similarity reward rather than only a sparse exact-match signal, and Figure~\ref{fig:reward-curves} shows that the IUPAC-related transformation rewards do increase over the course of training. The natural interpretation is that Aspen is learning meaningful partial regularities in nomenclature---getting more fragments and substructures of the name right---without yet mastering the full global syntax needed for exact correctness. Evaluated under the strict exact-equivalence criterion of Figure~\ref{fig:pairedLollipop}, those partial gains are largely invisible. In that light, Opus~4.6 is particularly impressive: it reaches 0.55 on IUPAC$\rightarrow$SMILES, 0.40 on SMILES$\rightarrow$IUPAC, 0.42 on InChI$\rightarrow$SMILES, and 0.18 on SELFIES$\rightarrow$SMILES. Given that the task examples are drawn from ZINC and often involve very long, intricate IUPAC and InChI strings, that level of exact performance is striking.

\paragraph{Multiproperty-constrained generation task.}

The multiproperty-constrained generation task is arguably the most application-relevant single-turn environment, because it asks the model to generate a molecule in SMILES form that satisfies several simultaneous design constraints rather than merely answer a question about an existing molecule (or selecting from a set of molecules). This is much closer to real lead optimization, where one typically has to design under multiple, potentially competing requirements. On this task, Aspen’s improvement is especially important: compared with Qwen, the valid response rate rises from 0.77 to 1.00, individual-constraint satisfaction rate from 0.59 to 0.70, and all-constraint satisfaction rate from 0.09 to 0.21. On the hardest metric---satisfying all constraints simultaneously---Aspen slightly exceeds the frontier baselines in the figure (0.19--0.20 for GPT-5, GPT-5.2, Opus~4.0, and Opus~4.6). At the same time, the persistent gap between individual-constraint and all-constraint success across every model shows that constraint composition remains the core difficulty: generating a valid molecule is no longer the main bottleneck, but jointly satisfying several medicinal-chemistry requirements still is. From the perspective of practical drug discovery, however, this is one of the most encouraging results in Figure~\ref{fig:pairedLollipop}, because the largest within-family jump appears on the task that most directly resembles real molecular design.

\paragraph{Other tasks.}

The final task row shows a split between near-saturated structural recognition and still-difficult higher-order chemical reasoning. Substructure classification is strong across all families, and Aspen closes most of the gap to frontier models by improving from 0.88 to 0.96, making it competitive with the 0.95--0.97 range achieved by the closed models. In contrast, reaction outcome prediction and MCS identification remain challenging for all models. Aspen improves over Qwen on both, but the strongest performance comes from Claude Opus~4.6, which substantially outperforms its 4.0 predecessor on reaction outcome prediction (0.31$\rightarrow$0.51) and MCS identification (0.00$\rightarrow$0.12). Thus, while simple structural recognition is now largely reliable, complex graph and reaction reasoning remain meaningful open challenges in the single-turn setting.

\subsection{Simulated lead-optimization}
\label{sec:lead-optimization}

\begin{figure}[htb]
    \centering
    \includegraphics[width=\linewidth]{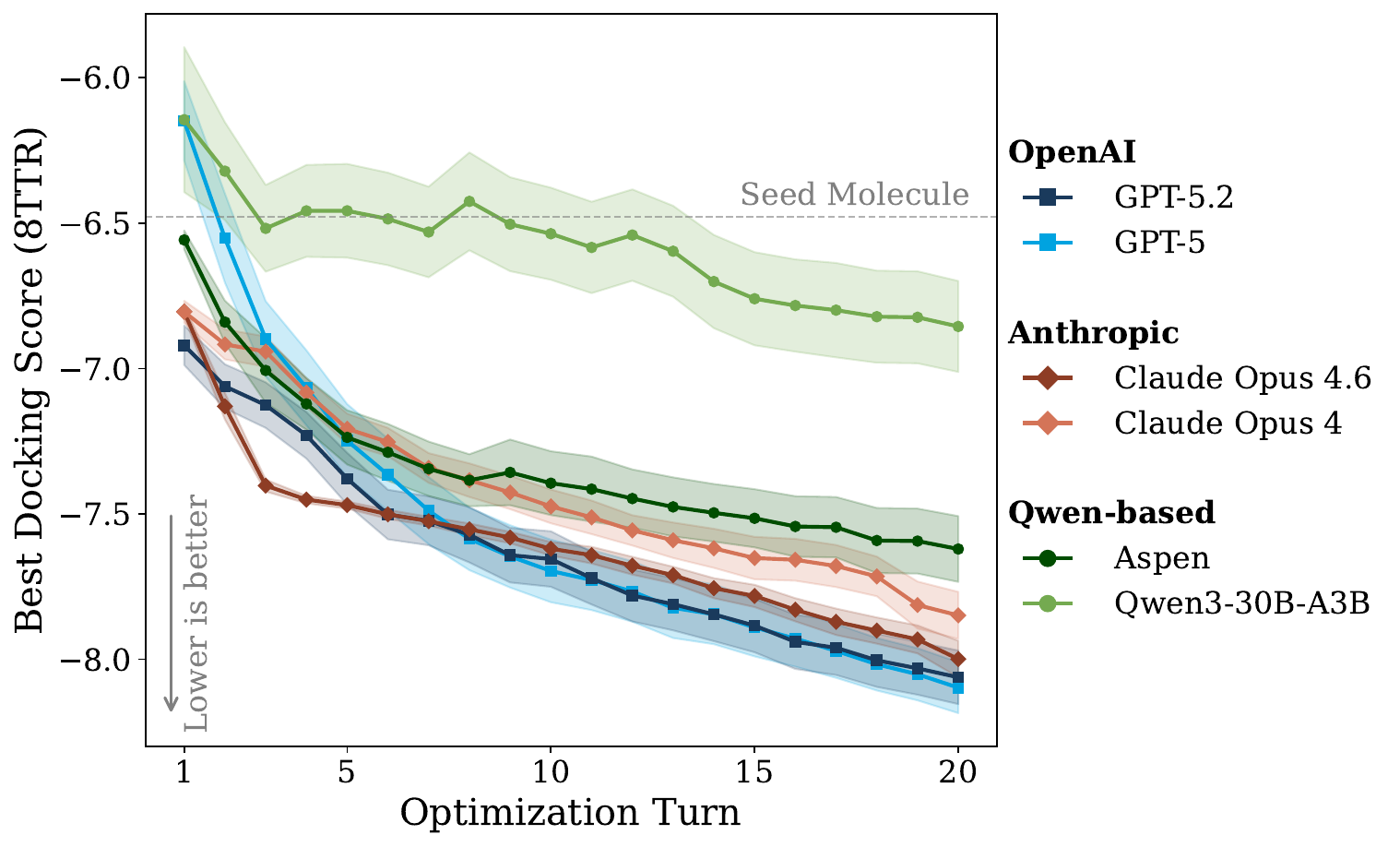}
    \caption{Mean best docking score over 20 optimization turns for 8TTR, averaged across 30 independent trajectories per model with shaded bands indicating standard error. Across all three model families, later versions outperform earlier ones in both final docking score and optimization efficiency. The improvement is most pronounced between the base Qwen model and Aspen, where the base model struggles to improve beyond the seed molecule while Aspen rapidly finds molecules with substantially lower docking scores. GPT-5 and GPT-5.2 converge to similar final scores but GPT-5.2 reaches them more efficiently, and Opus~4.6 consistently outperforms Opus~4 throughout the trajectory.}
    \label{fig:optimization-dynamics}
\end{figure}

Next, we examine the capabilities of these models in a multi-turn setting, consisting of an iterative molecular optimization loop that simulates real-world lead optimization.
Specifically, we consider optimizing a docking score---as a weak proxy for potency---under a set of property constraints, consisting of a combination of DMPK and RDKit properties that can all be computed. 
For the experiments here, we consider a single target, a carbonic anhydrase IX (PDB ID: \texttt{8TTR}), and a starting seed molecule obtained from~\citep{elshamsy2025design-e9e}. 
Each trajectory starts from the same molecule and runs for 20 turns; at each turn the model proposes a modified SMILES and is provided with the resulting docking score and the corresponding property values of the constraints (see Appendix~\ref{app:docking}). 
In the system prompt (see Appendix~\ref{app:single_turn}), we instruct the model to always propose structurally novel molecules across all 20 turns. Importantly, we blind the target name from the model, resulting in a black-box optimization setting that restricts the model from leveraging knowledge about the protein to carry out the optimization. 
We evaluate all three model families across 30 independent trajectories per model.

Across all three model families, later versions consistently outperform earlier ones (see Fig~\ref{fig:optimization-dynamics}, both in terms of raw docking score and optimization efficiency. 
This improvement is the most pronounced for the Qwen family of models; the base Qwen model notably struggles, both in generating valid molecules and generating molecules with low docking scores. 
After post-training, Aspen substantially improves on the base model, generating mostly valid molecules with docking scores well below that of the seed molecule. 
Within the OpenAI family of models, GPT-5.2 is able to more efficiently optimize the molecule in comparison to GPT-5, although both converge around a similar mean docking score after turn 10. 
Similarly, we observe that Opus~4.6 is better than Opus~4 at this design task, both in optimization efficiency and docking score.

\begin{figure}[htb]
    \centering
    \includegraphics[width=\textwidth]{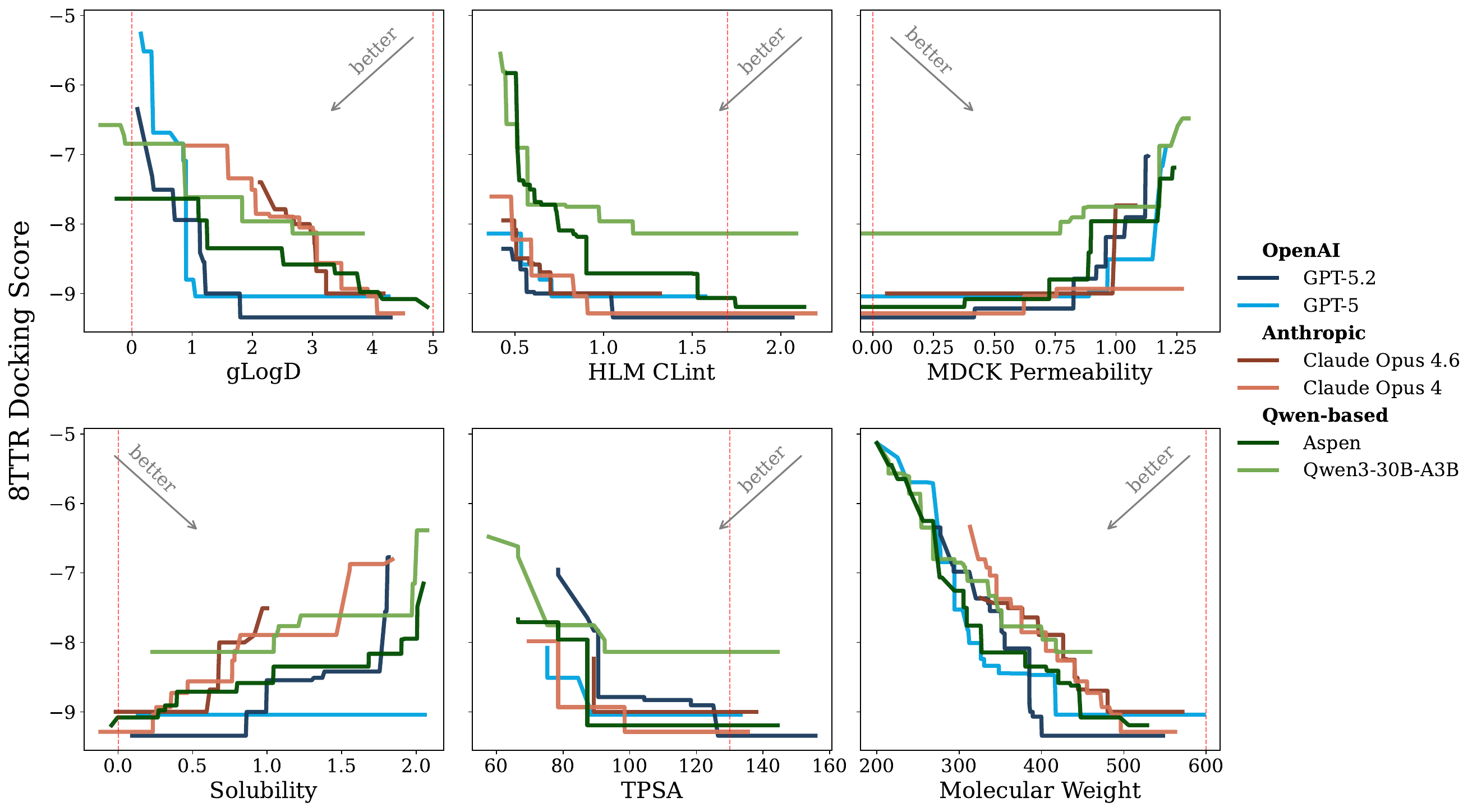}
    \caption{Pareto tradeoffs between docking score and molecular property constraints across models. Each panel plots docking score against one constraint, providing a two-dimensional view of the multi-objective optimization problem. The dashed red lines denote the constraints provided to the model. Points represent generated molecules across all turns and trajectories. The distributions highlight how models balance potency and constraint satisfaction; notably, Aspen achieves larger improvements in docking score without corresponding increases in size, consistent with its higher ligand efficiency relative to other frontier models.}
    \label{fig:pareto-frontier}
\end{figure}

Importantly, the docking score optimization involves adhering to certain constraints, which often introduces tradeoffs. 
For example, docking scores are loosely correlated with molecular weight; larger molecules can make more contacts, and thus increasing molecular size can improve the docking score (i.e.,\ making it more negative). 
In Figure~\ref{fig:pareto-frontier}, we present the tradeoffs between the six constraints and the docking score. 
We note that some of these constraints are less correlated with docking score than others, and that this 
represents only a two-dimensional slice of a higher-dimensional optimization problem.
Nevertheless, these plots help understand how each model navigates trade off between various constraints.

\begin{figure}[htb]
    \centering
    \includegraphics[width=\textwidth]{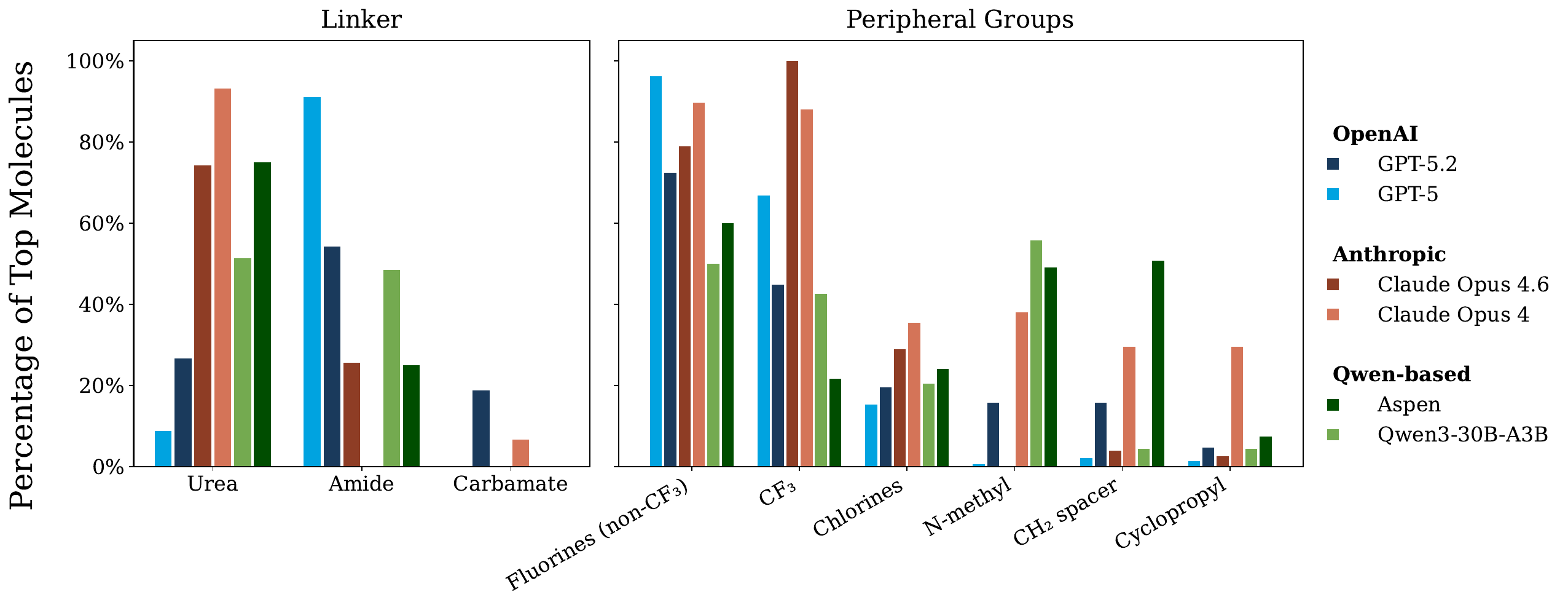}
    \caption{Chemical strategies in the top 25\% of scaffold-matching molecules. \textbf{Left:} Fraction of molecules retaining the seed's urea linker vs.\ converting to amide or carbamate. GPT-5 overwhelmingly converts to amide, while Aspen and Opus~4.6 more often retain the urea. \textbf{Right:} Mean count of key substructural features per molecule. Frontier models favor fluorination (including \ce{CF3}, particularly in the Opus family). In contrast, Aspen and Opus~4 show increased use of \ce{CH2} spacers and N-methylation, while Opus~4 is additionally enriched in more rigid groups such as cyclopropyl.}
    \label{fig:chemical-strategy}
\end{figure}

We find that models in the Qwen family generally struggle to generate molecules that simultaneously achieve low Human Liver Microsome (HLM) intrinsic clearance (CLint) and good docking scores, as reflected in the the constraint satisfaction rates (Figure~\ref{fig:constraint-satisfaction}). As conservative modifications are exhausted early on, Aspen increasingly introduces \ce{CH2} spacers and N-methylation (Figure~\ref{fig:chemical-strategy}), resulting in more flexible and lipophilic molecules associated with higher metabolic clearance and reduced HLM stability as it pursues additional potency. A similar, though less pronounced, trend is observed in Opus~4, which also shows lower HLM CLint satisfaction relative to other frontier models. As shown in Figure~\ref{fig:chemical-strategy}, Opus~4 also exhibits an increased use of \ce{CH2} spacers and N-methylation, although at a lower frequency than Aspen.
Despite this, Aspen is able to generate molecules with high ligand efficiency (i.e., the rate at which the docking score can be improved while minimizing size increases), especially in comparison to other frontier models.
In contrast, we find that the base Qwen model is generally the most Pareto-inefficient across all properties.

Finally, we examine the structural modifications made by each model to the starting scaffold, namely the central urea linker in the starting scaffold.
For top-scoring molecules, GPT-5 almost exclusively converts this linker to an amide, where the two Claude Opus models and Aspen generally maintain this linker as a urea and focus more on modifying peripheral R groups.
In their peripheral modifications, the Claude Opus models favor fluorination, especially in comparison to the Qwen family of models.
We additionally examine the fraction of unique molecules across all trajectories. We see that Opus~4.6 produces a lower fraction of unique molecules compared to all other models (Figure~\ref{fig:chemical-diversity}), indicating a narrower set of generated compounds. 
This is in stark contrast to the older Claude Opus~4 model, which has a fraction of unique molecules comparable to the other four models, likely hinting at some form of mode collapse in chemical space during the training from Claude Opus~4 to Claude Opus~4.6. 
\newline

Overall, Figures~\ref{fig:pairedLollipop}--\ref{fig:chemical-strategy} illustrate both the trainability and the current limitations of open-weight base models for specialized chemical tasks and more realistic multi-turn molecular optimization. \texttt{Qwen3-30B-A3B-Thinking-2507} starts substantially behind frontier closed models on many single-turn chemical tasks and also struggles in the simulated lead-optimization environment, but RL post-training closes much of that gap, rescues outright failure modes, and carries over to a much stronger multi-turn optimizer in Aspen. At the same time, the pattern across tasks suggests that RL is most effective when the base model already possesses at least partial chemical competence; on tasks where Qwen appears to have essentially no useful prior knowledge, reward shaping alone is not sufficient to reach strong performance. For such tasks, some combination of chemistry-specific supervised fine-tuning, continued midtraining on molecular and nomenclature-rich corpora, and broader exposure to the relevant representation languages is likely needed to provide the initial substrate on which RL can effectively build. The closed-model comparisons further show that later model versions continue to improve on many of the most chemically meaningful tasks in both the single-turn and multi-turn settings, even if that progression is not perfectly monotonic within every family. Claude Opus~4.6 is the strongest model on many of the hardest transformation and experimental-prediction tasks, while frontier closed models still set the pace on overall optimization efficiency in the lead-optimization loop; however, the largest within-family jump is Qwen-to-Aspen, indicating that careful post-training can move a smaller open model substantially closer to frontier behavior despite a much weaker starting point. More broadly, these results point toward a promising roadmap for future specialized molecular models: expanding the set of drug-discovery-specific environments and tasks, including more sophisticated environments that execute agentic workflows; leveraging proprietary internal data where available; and jointly scaling model size, task diversity, training duration, and chemistry-aware midtraining/SFT. Together, these directions may enable future models not only to answer chemical questions more accurately, but to function as more capable molecular design systems for real-world drug discovery.

\section{Conclusion}

In this work, we provide a framework for analyzing the progression of LLM capabilities for small-molecule drug design tasks. This is especially important given the ``jagged frontier'' nature of LLM capabilities. For open-weight models, we assess how easily their performance in small-molecule tasks can be improved with a simple post-training recipe. Additionally, we consider two families of closed frontier models and assess their improvements over time to better understand how closed models are improving for small-molecule chemistry tasks. We find that with post-training, we can significantly improve the performance of a 30B parameter model on these tasks. Furthermore, we see a clear improvement in the Anthropic family of models---due to deliberate training by the Anthropic team on chemistry tasks~\citep{anthropic2026claude-8ac}---while we see no significant gains for the OpenAI family of models on these tasks. Finally, while post-training enables meaningful gains across many tasks, we do not see any noticeable improvement from post-training on more difficult tasks, suggesting that we require further base-model training (e.g., midtraining) on these domains before RL-based post-training can be useful. Here, we exclusively focused on assessing the performance of the base model; however, truly understanding the promises and limitations of applying LLMs for drug discovery contexts will require understanding how LLM-based agents perform in real-world settings.

\section*{Acknowledgments}
We thank Hoang Nguyen, Charlie Gebhardt, Slaton Lipscomb, and David Konerding for their invaluable support in configuring the distributed training infrastructure used in this work.

\newpage
\FloatBarrier
\renewbibmacro{in:}{%
  \ifentrytype{article}
    {\iffieldundef{journaltitle}
       {}
       {\printtext{\bibstring{in}\intitlepunct}}}
    {\printtext{\bibstring{in}\intitlepunct}}}

\setlength{\bibitemsep}{0.5\baselineskip}
\printbibliography

\FloatBarrier

\newpage
\appendix

\section{Training Details}
\label{app:training_details}

We post-train \texttt{Qwen3-30B-A3B-Thinking-2507}~\citep{yang2025qwen3-84a} using the DAPO variant of GRPO~\citep{yu2025dapo-2df, shao2024deepseekmath-82a}, as implemented in the \texttt{trl} \texttt{GRPOTrainer} (v0.24.0)~\citep{vonwerra2020trl}. The training dataset comprises approximately 900k prompts---roughly 300k for the multiproperty constrained generation task and 600k distributed across all other tasks described in Appendix~\ref{app:task-details-appendix}. All tasks are mixed uniformly via dataset shuffling, without curriculum learning or task scheduling. We perform full-parameter fine-tuning in \texttt{BF16} mixed precision with Flash Attention~2~\citep{dao2023flashattention-2-89e}. No parameter-efficient methods (e.g., LoRA) are used. Algorithm~\ref{alg:grpo} summarizes the training procedure.

Training is distributed across 32 nodes, each equipped with 8 NVIDIA B200 GPUs (256 GPUs total), using DeepSpeed ZeRO Stage~3 for parameter, gradient, and optimizer sharding. Completions are generated using vLLM (v0.10.2)~\citep{kwon2023efficient} in colocate mode, where the inference engine shares GPU memory with the training process via an alternating sleep/wake cycle. We train for a single epoch (1{,}770 steps) with an effective batch size of 2{,}048 completions per step (512 unique prompts $\times$ 4 generations each) at a learning rate of \num{1e-6}. We train in a purely on-policy approach and with no KL-divergence penalty. Finally, we apply a soft overlong punishment that linearly penalizes completions between 16{,}384 and 32{,}768 tokens. Training took approximately 20 days. Table~\ref{tab:training_hyperparams} summarizes the key hyperparameters.

\begin{algorithm}[ht]
\caption{DAPO-based post-training}
\label{alg:grpo}
\begin{algorithmic}[1]
\Require Base policy $\pi_{\textrm{ref}}$; tasks $\{\mathcal{D}_i, R_i\}_{i=1}^T$; group size $G$; training steps $K$
\State Initialize $\pi_\theta \leftarrow \pi_{\textrm{ref}}$
\For{$k = 1, \dots, K$}
    \State $\pi_{\textrm{old}} \leftarrow \pi_\theta$
    \For{each task $i$ in minibatch}
        \State Sample $\boldsymbol{x} \sim \mathcal{D}_i$, then $\{\boldsymbol{y}_j\}_{j=1}^G \sim \pi_{\textrm{old}}(\cdot \mid \boldsymbol{x})$
        \State Compute $r_j = R_i(\boldsymbol{x}, \boldsymbol{y}_j)$ and $\hat{A}_j = (r_j - \mu_G) / \sigma_G$
    \EndFor
    \State $\theta \leftarrow \theta + \alpha \, \nabla_\theta \mathcal{J}(\theta)$ \Comment{Eq.~\ref{eq:grad_J}}
\EndFor
\end{algorithmic}
\end{algorithm}

\begin{table}[ht]
\centering
\small
\caption{Key training hyperparameters.}
\label{tab:training_hyperparams}
\begin{tabular}{lr}
\toprule
\textbf{Parameter} & \textbf{Value} \\
\midrule
Base model & \texttt{Qwen3-30B-A3B-Thinking-2507} \\
Total / active parameters & 30B / ${\sim}$3B (MoE) \\
Precision & BF16 \\
Learning rate & \num{1e-6} \\
Effective batch size & 2{,}048 \\
Unique prompts per step & 512 \\
Generations per prompt ($G$) & 4 \\
Training steps / epochs & 1{,}770 / 1 \\
$\epsilon_{\textrm{low}}$ / $\epsilon_{\textrm{high}}$ (clip bounds) & 0.2 / 0.28 \\
$\beta$ (KL coefficient) & 0.0 \\
Max completion length & 32{,}768 tokens \\
Overlong penalty onset & 16{,}384 tokens \\
Hardware & 256 GPUs (32 $\times$ 8 B200) \\
Parallelism & DeepSpeed ZeRO-3 + vLLM colocate \\
Wall-clock time & ${\sim}$20 days \\
\bottomrule
\end{tabular}
\end{table}

\section{Task Details}
\label{app:task-details-appendix}

\subsection{Reward definitions}
\label{sec:rewards}
Below we outline the reward functions used for training and evaluation.

\paragraph{Exponential-MSE reward}
For all numerical prediction tasks, we use the following reward
\begin{equation}
r = \exp\!\left(-k \cdot \frac{(\hat{y} - y)^2}{\sigma^2}\right),
\label{eq:exp-mse}
\end{equation}
where $\hat{y}$ is the predicted value, $y$ is the ground truth, $\sigma$ is the standard deviation---computed on the training data---of that property, and $k=5.0$ across all tasks. 
This reward lies in $(0, 1]$ and penalizes large errors quadratically before the exponential transform.

\paragraph{Binary equivalence reward}
Multiple-choice, substructure classification, and several other tasks use a binary equivalence reward,
\begin{equation}
r = \mathbf{1}[\hat{y} \equiv y],
\label{eq:binary-equivalence}
\end{equation}
which is a simple case-insensitive string match.

\paragraph{Dense equivalence reward}
For transformation tasks, we define the following dense reward
\begin{equation}
r = (1 - w)\,\mathbf{1}[\hat{y} \equiv y] + w \cdot \mathrm{sim}(\hat{y}, y).
\label{eq:equivalence}
\end{equation}
The first component of the reward is a simple string equivalence between the predicted and correct answer, and the second component measures how similar the predicted and correct answers are.
This provides richer information during training on how close the predicted molecules are to the ground-truth answer.
Tasks that output SMILES or require structural comparison (IUPAC$\rightarrow$SMILES, protomer, tautomer, Murcko scaffold, and reaction outcome prediction) use substructure-count similarity (Eq.~\ref{eq:sub-sim}).
SMILES$\rightarrow$IUPAC uses IUPAC semantic similarity (Eq.~\ref{eq:iupac-sim}), and SMILES$\rightarrow$formula uses normalized element-count similarity (Eq.~\ref{eq:formula-sim}). 
Tasks involving bidirectional representation conversion (e.g., IUPAC$\leftrightarrow$SMILES) use $w = 0.7$, placing greater weight on partial similarity to provide denser learning signal for these harder translations. 
Tasks involving structural modifications (tautomer, protomer, Murcko scaffold, formula) use $w = 0.3$, placing greater weight on exact equivalence.

\paragraph{Constraint satisfaction reward.}
Multi-property constrained generation uses a reward equal to the fraction of satisfied constraints,
\begin{equation}
r = \frac{1}{N}\sum_{i=1}^{N} \mathbf{1}[c_i \text{ satisfied}],
\label{eq:constraint-sat}
\end{equation}
where $N$ is the number of constraints and $c_i$ denotes the $i$-th property or scaffold constraint.

\paragraph{Formatting and length penalties.}
Across all tasks, an invalid completion penalty of $r = -0.2$ is assigned when the output is malformed, such as missing \texttt{<|answer\_start|>} tags or providing a reasoning summary shorter than 300 characters. An invalid output penalty of $r = -0.1$ is assigned when an answer can be extracted but does not match the required format (e.g., non-numeric text where a number is expected, or an invalid SMILES string). Additionally, a soft overlength penalty~\citep{yu2025dapo-2df} is applied to discourage unnecessarily long completions,
\begin{equation}
R_{\text{length}}(y) = \begin{cases}
0, & |y| \le L_{\max} - L_{\text{cache}} \\[4pt]
\dfrac{(L_{\max} - L_{\text{cache}}) - |y|}{L_{\text{cache}}}, & L_{\max} - L_{\text{cache}} < |y| \le L_{\max} \\[4pt]
-1, & |y| > L_{\max}
\end{cases}
\label{eq:overlong-penalty}
\end{equation}
where $|y|$ is the completion length in tokens, $L_{\max} = 32{,}768$ is the maximum completion length, and $L_{\text{cache}} = 16{,}384$ is the buffer over which the penalty is linearly interpolated from 0 to $-1$.

\subsection{Additional small-molecule task details}
\label{sec:task_specific_details}

Here we provide additional information about the small-molecule tasks.
\begin{itemize}
    \item \textbf{RDKit prediction.} Targets are computed with RDKit and rounded according to property-specific precision settings. Tanimoto similarity (used in the generation reward and tanimoto similarity task) is computed from Morgan fingerprints with radius 2 and 2048 bits, with chirality ignored.
    \item \textbf{Experimental prediction.} SAR and DMPK regression tasks use 5 in-context examples. Multiple-choice SAR and DMPK tasks use 15 in-context examples. Experimental tasks span internal assay potency, FS-Mol potency \citep{2021fs_mol}, and internal DMPK clearance, LogD, and solubility datasets.
    \item \textbf{Multiple choice.} Tasks include SMILES equivalence and multiple-choice SAR/DMPK settings in which the correct option is determined by structural identity or highest measured activity or endpoint value. For the SMILES equivalence task, the correct answer is a randomized SMILES representation of the query molecule, and the three incorrect choices are generated by converting the query SMILES to SELFIES~\citep{krenn2019self-referencing-91c} and applying random token-level perturbations (substitution, insertion, deletion, or swapping of SELFIES tokens) before converting back to SMILES. Because SELFIES guarantees chemical validity by construction, this procedure produces molecules that are always valid yet structurally distinct from the query. 80\% of the incorrect choices across all examples are generated such that they have the same molecular formula as the correct answer (excluding hydrogens) in order to force the model to reason about connectivity instead of simply counting atoms in the SMILES.
    \item \textbf{Transformation.} Evaluation combines format-specific validity with exact or structural equivalence, depending on the task. Protomer targets are defined at physiological pH (7.4), and protomer and tautomer tasks use balanced same/different sampling to avoid trivial copying.
    \item \textbf{Constrained generation.} Each example contains 1--5 regular constraints derived from a sampled ZINC molecule so that at least one valid solution exists. Constraints may include continuous property ranges, integer constraints, and BRICS-derived scaffold constraints. Additionally, up to three element-count constraints (not carbon or hydrogen) may be added.
    \item \textbf{Other.} This group contains three tasks:
    \begin{itemize}
        \item \textit{Substructure classification}: Given a molecule in SMILES and a functional group name, the model must answer ``yes'' or ``no'' as to whether the molecule contains that functional group. Ground-truth labels are assigned by SMARTS matching in RDKit using the \texttt{exmol} functional-group definitions \citep{exmol1,exmol2,exmol3}. The reward is binary equivalence (Eq.~\ref{eq:binary-equivalence}).
        \item \textit{Reaction outcome prediction}: Given a set of reactants and a reaction SMARTS template, the model must predict the product SMILES. Reaction templates are derived from Pistachio mappings~\citep[v.2025Q2]{NextMovePitsachio}. The reward uses structural equivalence with substructure-count similarity at $w = 0.3$ (Eq.~\ref{eq:equivalence}).
        \item \textit{MCS identification}: Given two molecules in SMILES, the model must return the SMILES of their Maximum Common Substructure (MCS). The MCS is defined with strict atom-type, aromaticity, ring-membership, and bond-order matching, and must contain at least two atoms. The reward is binary molecular equivalence (Eq.~\ref{eq:binary-equivalence}), where equivalence is determined by comparing canonical SMILES after normalization.
    \end{itemize}
\end{itemize}

\subsection{Similarity and equivalence function details}
\label{sec:similarity_details}

This section provides full definitions of the similarity and equivalence functions used in the reward computations described above.

\paragraph{Molecular equivalence.}
Two molecular representations are considered equivalent if they resolve to the same canonical SMILES after normalization. For most tasks, normalization consists of salt removal (retaining only the largest fragment) followed by tautomer canonicalization via OpenEye's \texttt{OEQuacPac}. The tautomer prediction task uses a lighter normalization that applies salt removal only.

\paragraph{Substructure-count similarity.}
Several transformation tasks use substructure-count similarity as the partial-credit component. For each molecule, a feature vector is constructed by counting occurrences of 1{,}242 predefined SMARTS patterns. The similarity between two molecules is then computed as the Tanimoto coefficient over these count vectors:
\begin{equation}
\mathrm{sim}_{\mathrm{sub}}(m_1, m_2) = \frac{\sum_i \min(v_{1,i},\, v_{2,i})}{\sum_i v_{1,i} + \sum_i v_{2,i} - \sum_i \min(v_{1,i},\, v_{2,i})},
\label{eq:sub-sim}
\end{equation}
where $v_{1,i}$ and $v_{2,i}$ are the substructure match counts for pattern $i$.

\paragraph{Morgan fingerprint Tanimoto similarity.}
The Tanimoto similarity estimation task use standard Tanimoto similarity computed from Morgan (circular) fingerprints with radius 2, 2{,}048 bits, and chirality disabled (RDKit defaults).

\paragraph{Formula similarity.}
The SMILES$\rightarrow$formula task uses a normalized element-count similarity based on the L1 (Manhattan) distance between atom-count vectors:
\begin{equation}
\mathrm{sim}_{\mathrm{formula}}(f_1, f_2) = \max\!\left(0,\; 1 - \frac{\sum_{e} |c_{1,e} - c_{2,e}|}{\sum_{e} c_{1,e} + \sum_{e} c_{2,e}}\right),
\label{eq:formula-sim}
\end{equation}
where $c_{1,e}$ and $c_{2,e}$ are the counts of element $e$ in each formula, and the sum is over all elements present in either formula.

\paragraph{IUPAC semantic similarity.}
The SMILES$\rightarrow$IUPAC task uses a multi-component semantic similarity metric that operates directly on IUPAC name strings without requiring successful chemical parsing. Each name is decomposed into seven semantic features, and the final score is a weighted combination modulated by a coverage factor:
\begin{equation}
\mathrm{sim}_{\mathrm{IUPAC}} = \gamma \cdot \sum_{j \in \mathcal{A}} \tilde{w}_j \, s_j,
\label{eq:iupac-sim}
\end{equation}
where $s_j$ is the score for component $j$, $\mathcal{A}$ is the set of components present in the reference name, $\tilde{w}_j = w_j / \sum_{k \in \mathcal{A}} w_k$ are dynamically renormalized weights, and $\gamma$ is a key-token coverage factor. The coverage factor $\gamma$ is the fraction of applicable key components (suffix, parent structure, substituents/unsaturation) from the reference that are also present in the prediction. The default component weights are:

\begin{center}
\begin{tabular}{lcl}
\toprule
\textbf{Component} & \textbf{Weight} & \textbf{Description} \\
\midrule
Suffix & 0.25 & Principal characteristic group match \\
Substituents & 0.20 & Substituent family F1 score \\
Parent & 0.18 & Parent chain length and ring topology \\
Locants & 0.13 & Positional agreement (Levenshtein + F1) \\
Stereochemistry & 0.12 & R/S and E/Z descriptor coverage and correctness \\
Unsaturation & 0.09 & Unsaturation morpheme (ene/yne) agreement \\
Trigram & 0.03 & Character trigram fallback \\
\bottomrule
\end{tabular}
\end{center}

\paragraph{Constraint types.}
Multi-property constrained generation supports three constraint types:
\begin{itemize}
    \item \textbf{Range}: $v_{\min} \leq \hat{v} \leq v_{\max}$, where either bound may be omitted (unbounded).
    \item \textbf{Exact}: $\hat{v} = v_{\mathrm{target}}$ (integer or categorical match).
    \item \textbf{Scaffold}: the generated molecule must contain the specified scaffold as a substructure (RDKit \texttt{HasSubstructMatch}).
\end{itemize}

\section{Environment Details}
\label{app:environment_details}

\subsection{Single-Turn Environment}
\label{app:single_turn}

\begin{lstlisting}[nolol, escapechar=¤]
 You are a chemistry expert.

  ¤\tprompt{IMPORTANT:}¤ Do not use ANY external tools. Make predictions based on chemistry knowledge and reasoning alone.

  Use standard units when applicable:
  - g/mol
  - Angstroms

  At the end of your response, provide a summary of your reasoning and the answer:
  On a new line provide the reasoning summary between ¤\treason{<|reasoning\_summary\_start|>}¤ and ¤\treason{<|reasoning\_summary\_end|>}¤ tags.
  On a new line provide the final answer between ¤\tanswer{<|answer\_start|>}¤ and ¤\tanswer{<|answer\_end|>}¤ tags.

  Example format:

  ¤\treason{<|reasoning\_summary\_start|>}¤[summary of reasoning and analysis]¤\treason{<|reasoning\_summary\_end|>}¤

  ¤\tanswer{<|answer\_start|>}¤[final answer]¤\tanswer{<|answer\_end|>}¤

  Note: The brackets [ ] above are placeholders only - do not include them in your actual response.

  Strict answer requirements:
  - Output exactly one reasoning summary between ¤\treason{<|reasoning\_summary\_start|>}¤ and ¤\treason{<|reasoning\_summary\_end|>}¤.
  - Output exactly one final answer between ¤\tanswer{<|answer\_start|>}¤ and ¤\tanswer{<|answer\_end|>}¤.
  - Inside the answer tags, include only the final value: no prose, no ranges, no qualifiers.
  - Do not guess randomly. Use your chemistry knowledge to reason and provide your single best chemically-informed prediction.
  - For numeric answers, output a single number with no units inside the answer tags unless explicitly requested.
  - For string answers, output only the string inside the answer tags.
  - Do not include any additional text on the answer line beyond the answer tags.
\end{lstlisting}
\captionof{lstlisting}{Core system prompt shared across all single-turn training and evaluation tasks. The prompt instructs the model to rely solely on internalized chemistry knowledge, produce a structured reasoning summary, and return a single final answer in a standardized format. This structured output enables automated reward computation during training and consistent evaluation across all tasks.}

\begin{lstlisting}[nolol, escapechar=¤]
¤\tprompt{USER PROMPT:}¤
What protonation state would CO[C@@H](CCNC(=O)[C@]1(NC(=O)OC(C)(C)C)C[C@H]2CCCC[C@H]21)C(=O)O assume in cellular conditions?
¤\tthink{<think>}¤
[internal model CoT hidden for readability]
¤\tthink{</think>}¤
¤\treason{<|reasoning\_summary\_start|>}¤
The molecule contains a terminal carboxylic acid group (C(=O)O) and amide functional groups. At cellular pH (~7.4), carboxylic acids (pKa ~4-5) are deprotonated, while amide groups (pKa ~15-20 for protons) and ester groups remain neutral. The carboxylic acid deprotonates to carboxylate (C(=O)[O-]), and no other ionizable groups (e.g., basic amines) are present. Thus, the protonation state in cellular conditions is the deprotonated form of the carboxylic acid.
¤\treason{<|reasoning\_summary\_end|>}¤
¤\tanswer{<|answer\_start|>}¤
CO[C@@H](CCNC(=O)[C@]1(NC(=O)OC(C)(C)C)C[C@H]2CCCC[C@H]21)C(=O)[O-]
¤\tanswer{<|answer\_end|>}¤
\end{lstlisting}
\captionof{lstlisting}{Example completion on a protonation state prediction task. The model first identifies the functional groups present in the molecule, a skill developed through the substructure classification tasks, before reasoning about their pH-dependent ionization to arrive at the correct protomer. This suggests that multi-task training encourages the composition of skills across task boundaries, where simpler capabilities like functional group identification serve as building blocks for more complex chemical reasoning.}

\begin{lstlisting}[nolol, escapechar=¤]
¤\tprompt{USER PROMPT:}¤
I need a molecule with:
- 1 to 3 total rings
- 0 to 2 NH or OH groups
- 2 aromatic rings
- 0 amide bonds
Suggest a SMILES string.
¤\tthink{<think>}¤
[internal model CoT hidden for readability]
¤\tthink{</think>}¤
¤\treason{<|reasoning\_summary\_start|>}¤
The molecule must have 1-3 total rings, 0-2 NH or OH groups, exactly 2 aromatic rings, and no amide bonds. Biphenyl (C1=CC=CC=C1-C1=CC=CC=C1) meets all criteria: it contains two aromatic rings (separate benzene rings), total rings = 2, no NH/OH groups (0), and no amide bonds. The SMILES string for biphenyl using standard aromatic notation (lowercase 'c' for aromatic carbons) is `c1ccccc1c2ccccc2`, which satisfies all conditions.
¤\treason{<|reasoning\_summary\_end|>}¤
¤\tanswer{<|answer\_start|>}¤
c1ccccc1c2ccccc2
¤\tanswer{<|answer\_end|>}¤
\end{lstlisting}
\captionof{lstlisting}{Example completion on a multiproperty constrained generation task. After proposing biphenyl as a candidate molecule, the model systematically verifies each property constraint using internalized knowledge of molecular descriptors, echoing the RDKit property prediction tasks. The model also explicitly reasons about aromatic SMILES notation and the use of lowercase \texttt{c} for aromatic carbons, reflecting familiarity with molecular representation conventions reinforced by the transformation tasks.}

\begin{lstlisting}[nolol, escapechar=¤]
¤\tprompt{USER PROMPT:}¤
What's the molecular formula that matches COCCNC(=O)CSc1ccccc1C(=O)N1CC[C@@H](C)O[C@@H](C)C1?
¤\tthink{<think>}¤
[internal model CoT hidden for readability]
¤\tthink{</think>}¤
¤\treason{<|reasoning\_summary\_start|>}¤
The provided SMILES string is parsed to determine the molecular formula by counting all atoms and calculating the hydrogen count using degrees of unsaturation (DU). The SMILES string is broken down as follows:

- **Carbon atoms**: The string contains 19 carbon atoms (5 from the first chain, 6 from the aromatic benzene ring, 1 from the second carbonyl, 2 from "CC", 2 + 2 from the two "[C@@H](C)" groups, and 1 closing the ring).
- **Oxygen atoms**: 4 oxygen atoms (from ether, two carbonyls, and one ether in the ring).
- **Nitrogen atoms**: 2 nitrogen atoms (from the amide and ring).
- **Sulfur atom**: 1 sulfur atom.
- **Hydrogen count**: Using DU = 7 (4 from benzene ring, 2 from carbonyl groups, 1 from the ring closure), the formula for hydrogen is derived as H = 2C + 2 + N - 2 * DU = 2(19) + 2 + 2 - 14 = 28.

The molecular formula is confirmed to be C19H28N2O4S.
¤\treason{<|reasoning\_summary\_end|>}¤
¤\tanswer{<|answer\_start|>}¤
C19H28N2O4S
¤\tanswer{<|answer\_end|>}¤
\end{lstlisting}
\captionof{lstlisting}{Example completion on a SMILES-to-formula transformation task. The model first performs systematic SMILES parsing to count all heavy atoms, then independently cross-checks the hydrogen count using a degrees-of-unsaturation (DU) calculation from physical chemistry (DU = 7, yielding H = 28). This composition of structural parsing with a physical chemistry formula illustrates how knowledge from the property prediction tasks can serve as a verification mechanism during transformation tasks, arriving at the correct formula C\textsubscript{19}H\textsubscript{28}N\textsubscript{2}O\textsubscript{4}S.}

\begin{lstlisting}[nolol, escapechar=¤]
¤\tprompt{USER PROMPT:}¤
I'm analyzing SAR data with these measured potency values (higher values are better):
Molecule: COc1cccc(C(C)NC(=O)c2ccc(-c3ccncc3)cc2)c1 -> 5.400
Molecule: O=C(NCc1ccccc1)c1ccc(-c2ccncc2)cc1 -> 5.000
Molecule: COc1cccc(C(C)NC(=O)c2ccc(-c3ccncc3)cc2)c1 -> 5.000
Molecule: COc1cccc(C(C)NC(=O)c2ccc(-c3ccncc3)c(C)c2)c1 -> 5.000
Molecule: COc1cccc(CNC(=O)c2ccc(-c3ccncc3)cc2)c1 -> 5.500
Molecule: COc1cccc(C(C)NC(=O)c2sc(-c3ccncc3)nc2C)c1 -> 5.500
Molecule: CCCOc1cccc(C(C)NC(=O)c2ccc(-c3ccncc3)cc2)c1 -> 5.300
Molecule: CCCOc1cccc(C(C)NC(=O)c2ccc(-c3ccncc3F)cc2)c1 -> 5.500
Molecule: [2H]C([2H])([2H])Oc1cccc(C(C)NC(=O)c2sc(-c3ccncc3)nc2C)c1 -> 5.500
Molecule: COc1cccc(C(C)NC(=O)c2ccc(-c3ccncn3)cc2)c1 -> 5.000
Molecule: CC(NC(=O)c1ccc(-c2ccncc2)cc1)c1ccccc1 -> 5.000
Molecule: O=C(NC(CO)c1ccccc1)c1ccc(-c2ccncc2)cc1 -> 5.200
Molecule: COc1cccc(C(C)NC(=O)c2ccc(-c3ccncc3C)cc2)c1 -> 5.100
Molecule: COc1cccc(C(C)NC(=O)c2cnc(-c3ccncc3)nc2)c1 -> 5.500
Molecule: O=C(NCc1ccc(Cl)cc1)c1cc2c(-c3ccccc3)[nH]nc2s1 -> 5.000
Which molecule has the highest potency?
A: COc1cccc(C(CN)NC(=O)c2ccc(-c3ccncc3)cc2)c1
B: COc1cccc(CNC(=O)c2cc3c(-c4ccccc4)[nH]nc3s2)c1
C: NCC(NC(=O)c1ccc(-c2ccncc2)cc1)c1ccccc1
D: O=C(NC(CO)c1cc(Cl)cc(Cl)c1)c1ccc(-c2ccncc2)cc1
¤\tthink{<think>}¤
[internal model CoT hidden for readability]
¤\tthink{</think>}¤
¤\treason{<|reasoning\_summary\_start|>}¤
To determine which molecule (A-D) has the highest predicted potency, the provided SAR data is analyzed for structural features correlating with high potency (values of 5.5 are the highest observed). Key patterns from the dataset:
- Potency of 5.5 is achieved with structural features including:
  1. **Simple alkyl chain** on the amide nitrogen (e.g., molecule 5: `CNC(=O)` instead of branched `C(C)NC`).
  2. **Heterocyclic substituents** on the amide-linked aromatic ring (e.g., molecule 6: thiazole, molecule 13: triazine-like).
- Molecules with branched chains (e.g., molecule 1) or aryl groups (e.g., molecule 2) generally exhibit lower potency (5.0-5.4).

Evaluating options:
- **Option A**: Contains a cyanide-functionalized branched chain (`C(CN)NC`). Uncommon substituent with no precedent in high-potency molecules; likely not optimal.
- **Option B**: Features a simple `CNC(=O)` chain and a heterocyclic ring containing sulfur and nitrogen (`[nH]nc3s2`), resembling molecule 6 (thiazole-based, potency 5.5). The benzothiazole-like structure is a known favorable motif for potency.
- **Option C**: Involves a longer chain (`NCC(N...)`) attached to an aryl group, similar to molecule 2 (potency 5.0), suggesting lower activity.
- **Option D**: Contains chlorinated aryl groups, akin to molecule 12 (potency 5.2); substituents are neutral or slightly detrimental to high potency.

Option B aligns with the most successful patterns (simple chain + heterocycle), matching the structural features of high-potency molecules (molecules 5, 6, 13). Its predicted potency is 5.5, matching the highest values in the dataset.
¤\treason{<|reasoning\_summary\_end|>}¤
¤\tanswer{<|answer\_start|>}¤
B
¤\tanswer{<|answer\_end|>}¤
\end{lstlisting}
\captionof{lstlisting}{Example completion on a multiple-choice SAR potency ranking task using public FS-Mol data. Given 15 in-context molecules with measured potencies, the model identifies two structural patterns correlated with high potency: simple alkyl chains on the amide nitrogen and heterocyclic ring substituents. It applies these patterns to rank the four candidates, selecting option B (a benzothiazole-containing molecule) over the correct answer C. Despite the incorrect prediction, the reasoning demonstrates meaningful SAR analysis, including functional group identification and analogy to in-context examples. The failure highlights the difficulty of extrapolating structure-activity relationships from limited data, where plausible medicinal chemistry reasoning can still lead to incorrect rankings.}

\begin{lstlisting}[nolol, escapechar=¤]
¤\tprompt{USER PROMPT:}¤
Considering SMILES: CN(CCc1cccc([N+](=O)[O-])c1)Cc1nc2c(s1)CCC2 | qed: 0.606, output a molecule with higher QED and strong similarity.
¤\tthink{<think>}¤
[internal model CoT hidden for readability]
¤\tthink{</think>}¤
¤\treason{<|reasoning\_summary\_start|>}¤
The goal is to increase the QED (Quantitative Estimate of Drug-likeness) of the given molecule while maintaining strong structural similarity. The original molecule has a QED of 0.606, with a nitro group ([N+](=O)[O-]) attached to an aromatic ring. Nitro groups are known to negatively impact QED due to their high polar surface area (PSA) and potential toxicity, as they often contribute to higher molecular weight and reduced drug-likeness. To improve QED, a minimal structural modification replacing the nitro group with a fluorine atom (F) is proposed. Fluorine is a common isosteric replacement that reduces molecular weight, lowers PSA, and maintains structural similarity (as both are substituents on the aromatic ring). The modified SMILES replaces [N+](=O)[O-] with F, resulting in a molecule with potentially higher QED due to improved physicochemical properties without altering the core scaffold significantly. This change is consistent with established medicinal chemistry strategies for optimizing drug-likeness.
¤\treason{<|reasoning\_summary\_end|>}¤
¤\tanswer{<|answer\_start|>}¤
CN(CCc1cccc(F)c1)Cc1nc2c(s1)CCC2
¤\tanswer{<|answer\_end|>}¤
\end{lstlisting}
\captionof{lstlisting}{Example completion on a QED optimization generation task. The model identifies the nitro group as the dominant liability, reasoning about its high PSA, molecular weight contribution, and potential for metabolic activation to toxic hydroxylamines. It proposes replacing the nitro group with fluorine, a well-established isosteric substitution in medicinal chemistry, producing a valid SMILES with 38\% QED improvement (0.606 $\rightarrow$ 0.838). This illustrates cross-task transfer of DMPK and toxicology knowledge into a molecular generation context.}

\begin{lstlisting}[nolol, escapechar=¤]
¤\tprompt{USER PROMPT:}¤
What's the SMILES for [C][O][C][=Branch1][C][=O][C@@H1][Branch1][=N][C][=C][C][=C][C][Branch1][Ring1][O][C][=C][Ring1][Branch2][N][Branch1][C][C][C][C][=C][C][=C][C][=C][C][=C][Ring1][=Branch1][N][=C][Ring1][#Branch2][O]?
¤\tthink{<think>}¤
[internal model CoT hidden for readability]
¤\tthink{</think>}¤
¤\treason{<|reasoning\_summary\_start|>}¤
The provided SMILES string contains non-standard elements such as `[Branch1]`, `[Ring1]`, and similar, which are not valid in standard SMILES notation. Standard SMILES uses parentheses `()` for branches and numbers for ring closures. After analyzing the structure, the non-standard tokens are replaced: `[Branch1]` becomes `(`, `[Branch2]` becomes `(`, and `[Ring1]` becomes `1` (with subsequent ring closures matching the numbers). Parentheses are balanced to ensure valid SMILES syntax, and the chiral notation `[C@@H1]` is interpreted as `[C@H]1` for proper stereochemistry representation. The corrected SMILES is constructed by replacing all erroneous tokens and ensuring valid branching and ring structures, resulting in a coherent molecule description.
¤\treason{<|reasoning\_summary\_end|>}¤
¤\tanswer{<|answer\_start|>}¤
COC(=O)C[C@H]1N=C(C=CC=C)C1OC2=CC=CC=C2C=CC=C(O)C
¤\tanswer{<|answer\_end|>}¤
\end{lstlisting}
\captionof{lstlisting}{Example failure on a SELFIES-to-SMILES transformation task. The input is a SELFIES string, a molecular representation language with tokens like \texttt{[Branch1]} and \texttt{[Ring1]}, but the model fails to recognize it as SELFIES entirely, instead treating the tokens as corrupted SMILES syntax and attempting to ``fix'' them. Although the model produces a valid SMILES string, it encodes the wrong molecule (the correct answer is \texttt{COC(=O)[C@@H](c1cccc(OC)c1)N(C)Cc1cc2ccccc2nc1O}). This failure illustrates a key limitation discussed in Section~\ref{sec:post-training}: RL post-training can sharpen existing knowledge but cannot inject understanding of representation systems absent from the base model's pretraining data.}

\FloatBarrier
\subsection{Simulated Lead-Optimization}
\label{app:docking}

\begin{figure}[!htb]
    \centering
    \includegraphics[width=\textwidth]{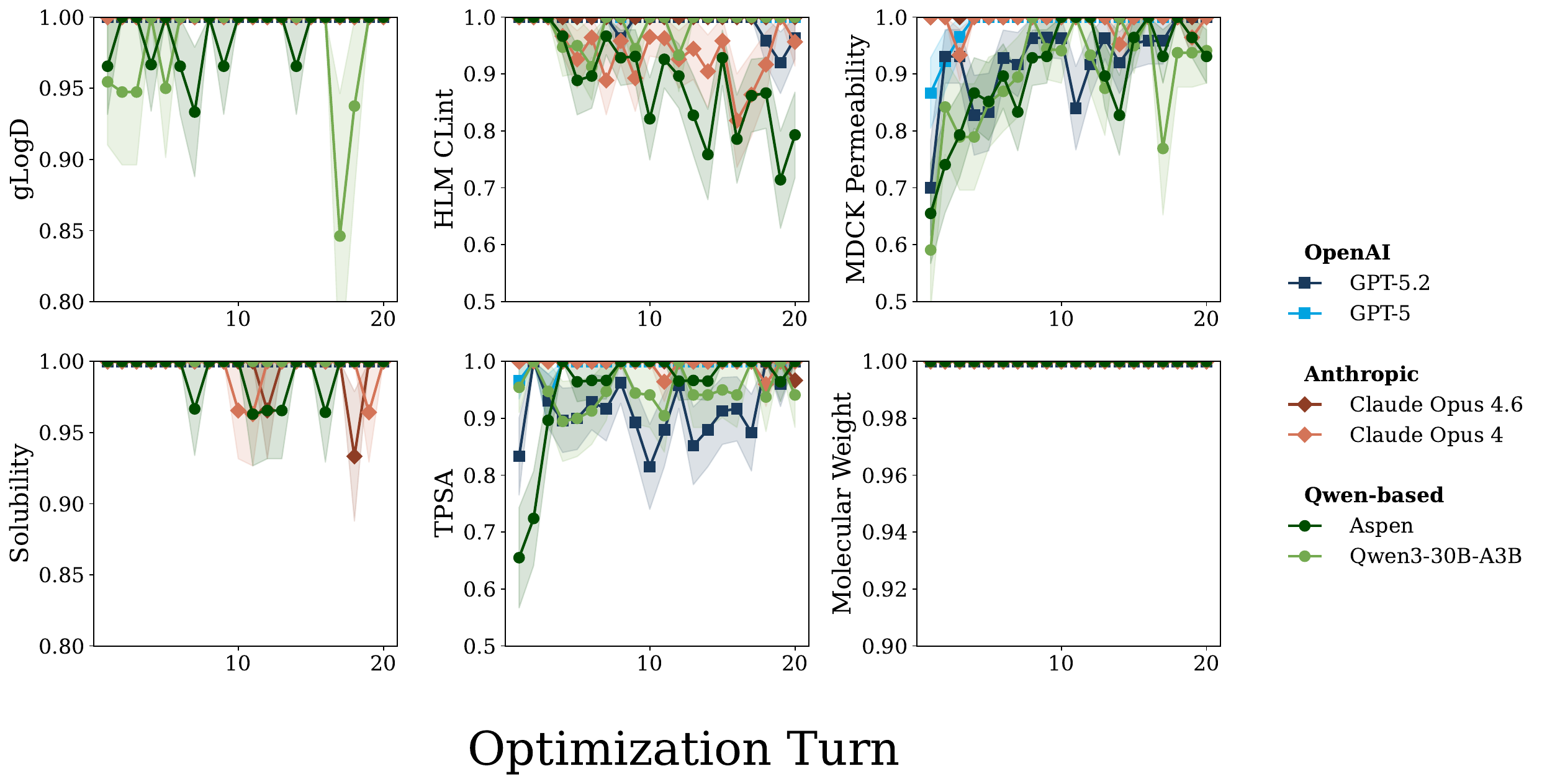}
    \caption{Constraint satisfaction rates over optimization turns (30 trajectories per model). Each panel shows the fraction of valid molecules satisfying the given DMPK constraint at each turn, with SEM bands. Aspen shows declining compliance on HLM CLint as trajectories progress, while frontier models maintain near-perfect satisfaction throughout.}
    \label{fig:constraint-satisfaction}
\end{figure}

\begin{figure}[!htb]
    \centering
    \includegraphics[width=\textwidth]{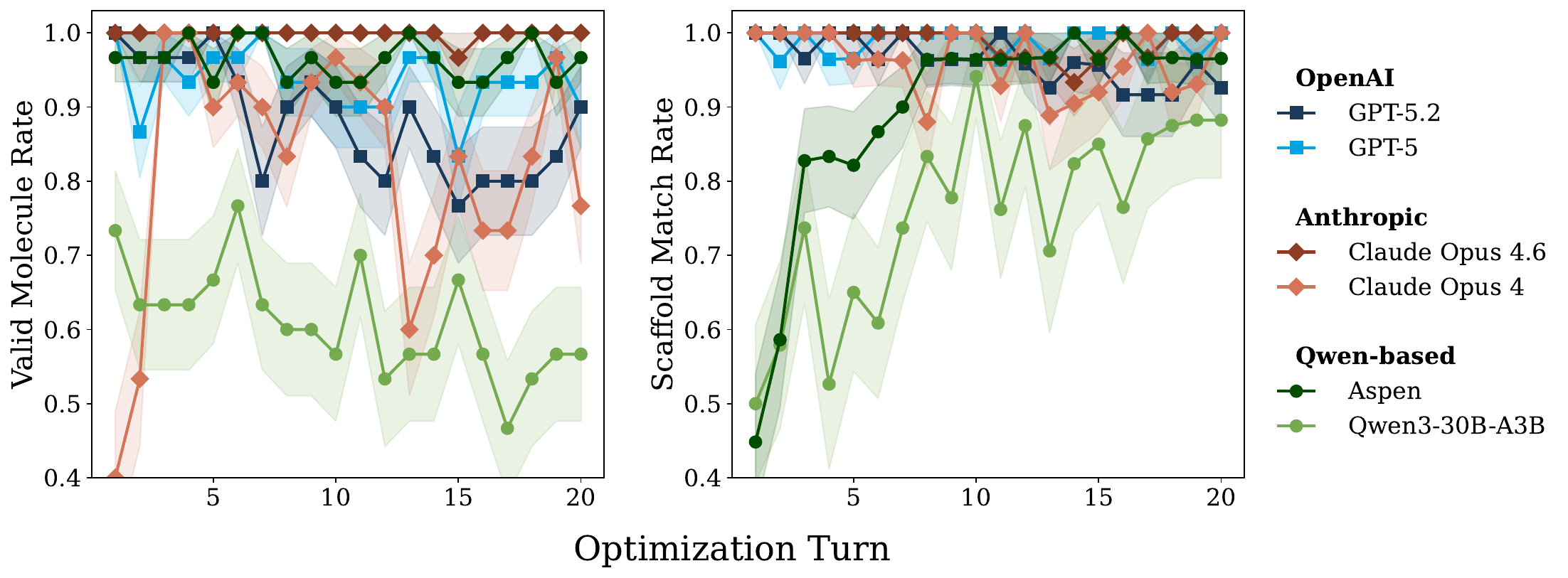}
    \caption{SMILES validity and scaffold retention over optimization turns (30 trajectories per model). \textbf{Left:} Fraction of turns producing a parseable SMILES. Opus~4.6 maintains near-perfect validity; Qwen3 base rarely exceeds 60\%. \textbf{Right:} Mean scaffold match rate among valid molecules. Qwen-based models start low but improve over turns, while frontier models stay above 90\% throughout.}
    \label{fig:validity-scaffold}
\end{figure}

All molecular optimization experiments are run with 20 turns with 30 total trajectories, starting from the same initial molecule: \texttt{NS(=O)(=O)c1ccc(NC(=O)Nc2ccc(F)cc2)cc1}. At each turn, the model generates a molecule in SMILES format, and tools are automatically called to compute the docking score and additional property scores. We computed the docking score for \texttt{8TTR} using the Vinardo~\citep{quiroga2016vinardo-2a7} scoring function as implemented in GNINA~\citep{mcnutt2021gnina-7f7}, with an exhaustiveness of 32 and an \SI{8.0}{\angstrom} buffer bounding box around the crystal ligand. We used \texttt{RDKit} functions and an internal DMPK oracle~\citep{napoli2025multitask-f0b} to predict all constraint properties. During initial experiments, we included the PDB ID (\texttt{8TTR}) in the system prompt; however, we observed that some frontier models made modifications based on an incorrect assumption about the target family. Nevertheless, this led to improved docking scores, possibly due to overlap with broadly favorable chemical features shared between the inferred protein family and the actual target. As a result, we blinded the target name in the system prompt to reflect a more black-box optimization setting. We plot the constraint satisfaction rate in Figure~\ref{fig:constraint-satisfaction}, the chemical validity rate and scaffold match rate in Figure~\ref{fig:validity-scaffold}, and the fraction of unique molecules in Figure~\ref{fig:chemical-diversity}. From these plots, we see that the post-trained model is able to generate a larger fraction of valid molecules that are generally diverse although it struggles to satisfy the HLM CLint constraint. Finally, we observe that the Opus 4.6 model struggles to generate unique molecules for this task.

\begin{lstlisting}[nolol, escapechar=¤]
You are a medicinal chemistry expert specializing in molecular optimization.

Your task is to design molecules that optimize potency while satisfying DMPK constraints.

After each turn, I will automatically calculate all properties for your proposed molecule and provide feedback.

[...instructions on formatting...]

¤\tprompt{================================================================}¤
¤\tprompt{CRITICAL: ALWAYS PROPOSE A NEW MOLECULE}¤
¤\tprompt{================================================================}¤

You MUST propose a structurally different molecule on EVERY turn. Never:
- Re-submit a molecule you have already proposed
- Declare that the current molecule is optimal or "good enough"
- Stop exploring modifications before all turns are exhausted

Even if the current molecule has excellent properties, there is always room for improvement. Continue exploring diverse structural modifications -- try different substituents, ring systems, linker variations, or stereochemistry changes. Your goal is to find the BEST possible molecule, not just a good one.
\end{lstlisting}
\captionof{lstlisting}{System prompt for the simulated lead-optimization environment (Section~\ref{sec:lead-optimization}). The prompt establishes a multi-turn interaction where the model proposes a modified molecule at each turn and receives automated property feedback. The novelty constraint at the bottom is critical for ensuring diverse exploration across the 20-turn trajectory, preventing the model from repeatedly proposing minor variants or declaring convergence prematurely.}

\begin{lstlisting}[nolol, escapechar=¤]
Optimize the following molecule for potency while satisfying these constraints:

¤\tlabel{Starting molecule:}¤ NS(=O)(=O)c1ccc(NC(=O)Nc2ccc(F)cc2)cc1

¤\tlabel{Constraints:}¤
- gLogD_value (lipophilicity): 0 to 5
- HLM_CLint_log (liver clearance): < 1.7 (corresponds to < 50 mL/min/kg)
- gMDCK_AB_log (permeability): > 0 (corresponds to > 1 x 10^-6 cm/s)
- LogKinSol (solubility): > 0 (corresponds to > 1 uM)
- tpsa: < 130 Angstroms^2
- molecular_weight: < 600 Da
- Must contain scaffold: O=CNc1ccc(S(N)(=O)=O)cc1

¤\tlabel{Objective:}¤
- Minimize docking score (lower is better)

¤\tlabel{Reference docking score:}¤ -6.48

Propose your first optimized molecule.
\end{lstlisting}
\captionof{lstlisting}{First-turn user prompt for the simulated lead-optimization environment. The prompt specifies the starting molecule, a scaffold substructure that must be preserved, six DMPK and physicochemical constraints with interpretable thresholds, and a docking score objective. On subsequent turns, this prompt is replaced by automated feedback containing the proposed molecule's computed property values and docking score.}

\begin{figure}[!htb]
    \centering
    \includegraphics[width=0.7\textwidth]{figs/chemical_diversity.pdf}
    \caption{Fraction of unique molecules proposed across all 30 trajectories per model in the simulated lead-optimization environment. Most models maintain a high fraction of unique molecules (0.86--0.95), but Claude Opus~4.6 is a notable outlier at 0.57, suggesting a degree of mode collapse in chemical space relative to its predecessor Opus~4 (0.88). This trend is consistent with the narrower structural strategies observed for Opus~4.6 in Figure~\ref{fig:chemical-strategy}.}
    \label{fig:chemical-diversity}
\end{figure}

\end{document}